\documentclass[%
    sigconf,
    preprint,
]{acmart}
\setcopyright{cc} 
\setcctype{by} 
\copyrightyear{2027}
\acmYear{2027}
\acmDOI{XXXXXXX.XXXXXXX}
\acmConference[KDD '27]{%
    ACM SIGKDD Conference on Knowledge Discovery and Data Mining
}{Month 00--00, 2027}{San Jose, CA, USA}
\acmBooktitle{%
    Proceedings of the 33rd ACM SIGKDD Conference on Knowledge Discovery and Data Mining (KDD '27)
}
\acmISBN{978-1-4503-XXXX-X/YYYY/MM}
\usepackage{bm}        
\usepackage{array}
\usepackage{multirow}  

\newcommand{\varphix}{\varphi(x)}             
\newcommand{\Eapp}[1]{\hyperref[#1]{SI~\ref{#1}}} 

\renewcommand\footnotetextcopyrightpermission[1]{} 
\begin{document}
\title[Covariance Last-Layer Ensembles]{Covariance Last-Layer Ensembles: Function-Space Diversity for Efficient Uncertainty Quantification}

\author{H. Martin Gillis}
\correspondingauthor
\email{martin.gillis@dal.ca}
\orcid{0000-0002-4650-5384}
\affiliation{%
 \institution{Dalhousie University}
 \city{Halifax}
 \state{Nova Scotia}
 \country{Canada}
}
\author{Isaac Xu}
\email{isaac.xu@dal.ca}
\orcid{0000-0003-4443-0582}
\affiliation{%
 \institution{Dalhousie University}
 \city{Halifax}
 \state{Nova Scotia}
 \country{Canada}
}
\author{Gabriel Spadon}
\email{spadon@dal.ca}
\orcid{0000-0001-8437-4349}
\affiliation{%
 \institution{Dalhousie University}
 \city{Halifax}
 \state{Nova Scotia}
 \country{Canada}
}
\author{Thomas Trappenberg}
\email{tt@cs.dal.ca}
\orcid{0000-0002-6144-8963}
\affiliation{%
 \institution{Dalhousie University}
 \city{Halifax}
 \state{Nova Scotia}
 \country{Canada}
}
\renewcommand{\shortauthors}{Gillis et al.}

\begin{abstract}
    A Last-Layer Ensemble (LLE), $K$ linear units on one shared frozen feature map, is an efficient single-pass approach to the disagreement-based epistemic uncertainty for out-of-distribution (OOD) detection.
    Its weakness is that members share the backbone gradient and can converge toward the same function, collapsing the inter-member diversity the signal depends on.
    Whether last-layer diversity can be restored, and what mitigates the collapse, is an open question.
    The weight-orthonormality defining Orthonormal Certificates (OC), the weight-orthonormal special case of the LLE, is only an indirect correction; it decorrelates the weights of the members, not their predictions.
    Here, we instead target the collapse directly in function space, with a Covariance Last-Layer Ensemble (cov-LLE) that places a direct covariance penalty on member activations.
    Cov-LLE restores the function-space diversity that weight-orthonormality cannot, and at matched $K$ recovers much of the diversity and calibration of a deep ensemble at $1\times$ backbone cost (in-distribution prediction variance $0.05\!\to\!9.3$ vs. $22.1$ ($\times10^{-3}$), and ECE $0.135\!\to\!0.090$ vs. $0.035$, for a $K\times$-cost deep ensemble), at no cost to accuracy.
    Viewing OC as a last-layer ensemble also organizes detectors into a two-axis taxonomy (by how their units are trained and how their outputs are scored) and exposes the OC score as a magnitude, motivating a scale-invariant, label-free direction score that repairs its near-OOD failure, adding $+0.16$ to $+0.18$ ROC AUC on every backbone.
\end{abstract}
%
%
%
%
\maketitle
\pagestyle{plain} 
%
\section{Introduction}
\label{sec:introduction}

A deployed neural network that cannot identify inputs outside its training distribution is a liability. 
The uncertainty-quantification literature separates two sources of uncertainty. 
Aleatoric uncertainty is the irreducible noise in the data, while epistemic uncertainty reflects what the model itself does not know, is high where training data was absent, and is the quantity relevant to Out-of-Distribution (OOD) detection, active learning, and safe deferral.
Epistemic uncertainty is not the same as low softmax confidence (networks are routinely confidently wrong on OOD inputs), which is why dedicated estimators are necessary.

A compute-efficient estimator places $K$ explicitly instantiated linear heads on one frozen feature map $\varphix$: a Last-Layer Ensemble (LLE), the explicit-members last-layer variant. 
This differs from placing a distribution over a single last layer (last-layer Laplace \citep{Kristiadi::2020aa,Daxberger::2021aa}, variational Bayesian last-layer \citep{Harrison::2024aa}, last-layer Monte Carlo dropout \citep{Gal::2016ab}).
One backbone pass then supplies $K$ low-cost heads whose disagreement estimates epistemic uncertainty, and Orthonormal Certificates (OC) \citep{Tagasovska::2019ab}, label-free detectors that run on the penultimate features of any pretrained network, are its weight-orthonormal special case.
However, a last-layer ensemble has a structural weakness in that its members share every gradient through the backbone, so they converge toward the same function and the inter-member disagreement that ensemble uncertainty depends on can collapse.
Whether last-layer diversity can nonetheless be restored well enough to recover the benefit of a full deep ensemble \citep{Lakshminarayanan::2017aa}, the $K$-network gold standard for epistemic uncertainty at $K\times$ the training and inference cost, and what mitigates this collapse, is an open question in efficient uncertainty quantification \citep{Kirsch::2025aa,Galdran::2023aa,Havasi::2021aa}.

The weight-orthonormality defining OC is already a collapse correction, but an indirect one. 
It decorrelates the certificates in weight space, which need not decorrelate their outputs in function space.
When a weight-orthonormality penalty is instead placed on an intermediate embedding rather than the output layer, the classifier negates the decorrelation almost entirely.
We therefore target the collapse where it matters most, directly in function space with a covariance penalty on member activations (cov-LLE).
Viewing OC as one corner of this design space also yields an efficiency result, a scale-invariant label-free score, and an organizing diagnosis of when the restored diversity improves detection.
This paper makes three main contributions:
\begin{enumerate}
    \item \textbf{A diagnostic account in which OC is a last-layer ensemble and feature quality decides whether it improves detection} (\hyperref[sec:natural]{Section~\ref{sec:natural}}). 
    A unifying $2\times2$ (member objective $\times$ scoring rule) places the canonical methods on its diagonal, and whether the recovered diversity converts into detection gains depends on feature quality, not bi-Lipschitz conditioning. 
    \item \textbf{Mitigating LLE collapse in function space (cov-LLE)} (\hyperref[sec:divobj]{Section~\ref{sec:divobj}}). 
    A covariance penalty on member activations restores the function-space diversity that weight-ortho\-normality cannot, recovering comparable diversity and calibration of a deep ensemble at $1\times$ cost.
    \item \textbf{A direction score that repairs the label-free near-OOD failure} (\hyperref[sec:direction]{Section~\ref{sec:direction}}).
    A scale-invariant, label-free direction score of the OC certificates improves near-OOD on every backbone and removes the SVHN inversion, showing the label-free failures of OC are scoring artifacts and providing an unsupervised OOD signal on any frozen backbone.
\end{enumerate}
\hyperref[sec:natural]{Section~\ref{sec:natural}} establishes the OC and LLE equivalency structure and the collapse on natural images, \hyperref[sec:divobj]{Section~\ref{sec:divobj}} introduces the function-space correction (cov-LLE), and \hyperref[sec:featquality]{Sections~\ref{sec:featquality}}--\ref{sec:generality} develop the feature-quality gate, the direction score, efficient scoring, the baseline comparison, and modality transfer.
Additional results are available in the Supporting Information (SI).

\section{Background and Related Work}
\label{sec:bg}

For a model $p(y\mid x)$, aleatoric uncertainty is the spread of $y$ at fixed $x$ and persists with infinite data; epistemic uncertainty reflects model ignorance and vanishes with infinite in-distribution (ID) data, and OOD detection is epistemic. 
The disagreement of a deep ensemble is commonly quantified by the mutual information between prediction and parameters (i.e., BALD \citep{Houlsby::2011aa})
\begin{equation}
    \label{eq:bald}
    \mathcal{I}(y;\theta\mid x)
    = \underbrace{H\!\left[\tfrac{1}{K}\textstyle\sum_k p_k(y\mid x)\right]}_{\text{total}}
    - \underbrace{\tfrac{1}{K}\textstyle\sum_k H\!\left[p_k(y\mid x)\right]}_{\text{aleatoric}},
    \end{equation}
where $p_k$ is the softmax of member $k$ and $H$ is Shannon entropy; the residual is the epistemic term, large only when members disagree. 
Last-layer ensembles freeze a shared backbone $\varphix$ and ensemble only the final linear head, so one backbone pass supplies $K$ low-cost heads.
However, heads can collapse and disagreement becomes uninformative, and vanilla LLEs (diversity only from random initialization) are prone to under-diversify, depending on how they are trained \citep{Lee::2015aa}. 
Efficient approximations exchange $K\times$ training strategies as a proxy for deep-ensemble diversity (BatchEnsemble \citep{Wen::2020aa}, MIMO \citep{Havasi::2021aa}, MHML \citep{Galdran::2023aa}, epinets \citep{Osband::2023aa}); the LLE is the limiting case, sharing the entire backbone and branching only at the head (TreeNets, \citep{Lee::2015aa}). 
A single-head family instead places a posterior over the last layer alone (last-layer Laplace \citep{Kristiadi::2020aa,Daxberger::2021aa}, variational Bayesian last layers \citep{Harrison::2024aa}, and last-layer Monte Carlo dropout \citep{Gal::2016ab}); the LLE is the explicit-members variant of the same scope. 
Our cov-LLE decorrelates member activations with a single Barlow-Twins / VICReg-style covariance penalty \citep{Zbontar::2021aa,Bardes::2022aa} (\hyperref[sec:divobj]{Section~\ref{sec:divobj}}).

\smallskip \noindent\textbf{Last-layer-ensemble collapse.}
\label{sec:collapsedef}
A last-layer ensemble collapses when its members converge to near-identical functions, so the prediction disagreement captured by the epistemic term of \autoref{eq:bald} vanishes. 
We quantify it in function space, as the across-member variance of the predicted distribution $\tfrac{1}{KC}\sum_{k,c}[p_k(c\mid x)-\bar p(c\mid x)]^2$ (over $K$ members and $C$ classes), or the magnitude-invariant $1-$linear CKA of the member activations \citep{Kornblith::2019aa}; full collapse is the limit where both vanish on ID and OOD inputs, leaving BALD uninformative. 
Since every member shares the backbone gradient through $\varphi$, a vanilla LLE (van-LLE) is prone to collapse, especially under extended training, a failure mode that \citet{Kirsch::2025aa} discusses. 
We distinguish this from \emph{feature collapse}, a property of the frozen backbone $\varphi$ no head diversity can undo, and from the rank degeneracy of OC as its certificate count $K$ approaches the feature dimension $d$ (\hyperref[app:divobj]{SI~\ref{app:divobj}}). 
Weight-orthonormality (OC) decorrelates the members in weight space, an \emph{indirect} mitigation; the function-space covariance penalty (cov-LLE) is a \emph{direct} one, acting on their activations.

\smallskip\noindent\textbf{Feature collapse and distance-aware representations.}
\label{sec:featcollapse}
OC \citep{Tagasovska::2019ab}, Mahalanobis \citep{Lee::2018ac}, and last-layer disagreement are all scores of feature geometry, so all inherit feature collapse, whereby a classifier-trained backbone may map far-OOD inputs onto the ID region of $\varphi$-space, destroying the signal before any detector sees it.
The distance-aware methods DUQ \citep{Amersfoort::2020aa}, SNGP \citep{Liu::2022ad}, and DUE \citep{Amersfoort::2021aa} make $\varphi$ approximately bi-Lipschitz
\begin{equation}
    \label{eq:bilip}
    L_{\text{low}}\,\lVert x-x'\rVert 
    \le 
    \lVert \varphi(x)-\varphi(x')\rVert 
    \le 
    L_{\text{high}}\,\lVert x-x'\rVert,
\end{equation}
with the upper bound (smoothness, against folding) from spectral normalization \citep{Miyato::2018aa} and the lower bound (sensitivity, against squashing) from residual connections; SNGP restricts spectral normalization at a coefficient $L$, where a larger $L$ exchange smoothing for the expressive power of the network.

\smallskip\noindent\textbf{Orthonormal certificates and feature geometry.}
OC \citep{Tagasovska::2019ab} learns $K$ linear certificates $\mathbb{C}\in\mathbb{R}^{d\times K}$ on frozen features $\varphix\in\mathbb{R}^{d}$ that annul ID data while remaining orthonormal
\begin{equation}
    \label{eq:oc}
    \min_{\mathbb{C}}\;\; 
    \mathbb{E}\!\left[\lVert \mathbb{C}^\top \varphix\rVert^2\right]
    \;+\; 
    \lambda_\mathrm{oc}\,\lVert \mathbb{C}^\top \mathbb{C} - I\rVert_F^2.
\end{equation}
The orthonormality term is essential; without it all the certificates collapse onto the lowest-variance direction or to zero. 
The test-time epistemic score is the certificate norm
\begin{equation}
    \label{eq:ocscore}
    \mathrm{S_N}(x) 
    = 
    \lVert \mathbb{C}^\top \varphix \rVert,
\end{equation}
small for ID and large for OOD inputs; it is a label-free analog of the Mahalanobis detector \citep{Lee::2018ac}.
We call this the OC norm, or its magnitude when contrasting it with the scale-invariant direction score (\hyperref[sec:direction]{Section~\ref{sec:direction}}).

\smallskip\noindent\textbf{Post-hoc OOD detectors.}
Most single-model OOD scores are post-hoc functions of the logits or features of a frozen network: 
(i) logit-space: MSP \citep{Hendrycks::2017aa}, 
ODIN \citep{Liang::2018aa},
energy \citep{Liu::2020ab}, 
ReAct \citep{Sun::2021aa}; 
(ii) feature-space: Mahalanobis \citep{Lee::2018ac}, 
KNN \citep{Sun::2022aa}, 
ViM \citep{Wang::2022aa}; 
and (iii) single-model epistemic estimators: Monte Carlo dropout \citep{Gal::2016ab}, 
DUQ / SNGP / DUE \citep{Amersfoort::2020aa,Liu::2022ad,Amersfoort::2021aa}, 
and OC \citep{Tagasovska::2019ab}. 
The LLEs are also post-hoc detectors, geometry- and disagreement-based scores of a frozen $\varphi$. 
For a broader discussion of the generalized-OOD setting, see \citet{Yang::2024aa}.

\section{Methods and Experimental Setup}
\label{sec:methods}
%
\noindent\textbf{Data, protocol, and backbones.}
We use a held-out-class OOD protocol: a backbone is trained on half the ID classes, the held-out classes form near-OOD, and a semantically distant dataset defines far-OOD.
Every detector operates on the same backbone, frozen unless otherwise noted, with our scores and the feature-space baselines running on its penultimate feature map $\varphix$ and the logit-space baselines on the classifier logits derived from it.
On natural images this is CIFAR-10 classes 0--4 as ID, 5--9 as near-OOD, and SVHN as far-OOD.
We evaluate on a 3-stage CNN (CIFARNet: three $3\times3$-conv stages of 32 / 64 / 128 channels with $2\times2$ max-pooling, a 256-d linear penultimate layer, and a 5-member head; accuracy: $0.821$), WideResNets (WRN-10-4 / 16-4 / 28-10, accuracy: $0.935$--$0.950$), and an OpenOOD ResNet-18 (accuracy: $0.940$).
We also use a pretrained 100-class CIFAR-100 VGG (accuracy: $0.740$) for scale-up, with CIFAR-100 as ID, CIFAR-10 and TinyImageNet as near-OOD, and SVHN, DTD, and MNIST as far-OOD (\hyperref[app:cifar100]{SI~\ref{app:cifar100}}).
For text, we use a frozen DistilBERT (accuracy: $0.872$) over 20 Newsgroups, with 5 ID and 5 held-out near-OOD categories and word-shuffled far-OOD (\hyperref[app:llm]{SI~\ref{app:llm}}).
A controlled MNIST setting (digits 0--4 / 5--9 / Fashion-MNIST) isolates each mechanism, using an MNISTNet CNN (two $3\times3$-conv layers of 16 / 32 channels with $2\times2$ max-pooling, a 128-d linear penultimate layer, and a 5-member head; \hyperref[app:divobj]{SI~\ref{app:divobj}}).

\smallskip\noindent\textbf{Orthonormal Certificates (OC).} $K=32$ certificates, $\lambda_\mathrm{oc}=5.0$ (OC detection is insensitive to the orthonormality strength for $\lambda_\mathrm{oc}>1$; \citep{Tagasovska::2019ab}), 1000 Adam steps at lr $10^{-2}$, objective \autoref{eq:oc} and score \autoref{eq:ocscore}.
Features are standardized using train statistics only.

\smallskip\noindent\textbf{Last-layer ensemble (LLE).} 
$K=32$ linear heads $\{W_k\in\mathbb{R}^{C\times d},b_k\}$ trained jointly by cross-entropy on the frozen features (Adam, lr $10^{-2}$, 40 epochs, batch size 512). 
Vanilla heads differ only by random initialization; ortho heads add an orthonormality penalty on their flattened, unit-normalized weight vectors 
\begin{equation} 
    \label{eq:ortho} 
    \Omega(W) 
    = 
    \frac{1}{K(K-1)}\sum_{i\neq j}\big(\hat{w}_i^\top \hat{w}_j\big)^2, 
    \qquad 
    \hat{w}_k 
    = 
    \frac{\mathrm{vec}(W_k)}{\lVert \mathrm{vec}(W_k)\rVert}, 
\end{equation}
as the mean squared cosine similarity between the weight vectors of distinct members. 
Since each vector is unit-normalized before the penalty is computed, the penalty constrains the direction of each vector but not the length, so it never restricts the magnitude of the head logits.
Penalty weight $\lambda_{\mathrm{ortho}}=3.0$ is used throughout. 
Unless otherwise noted, the LLE epistemic score is BALD \autoref{eq:bald} (the $\mathcal{O}(KC)$ mutual information); OC is scored by its certificate norm.

\smallskip\noindent\textbf{The unifying $\bm{2\times2}$.} 
We treat every detector as $K$ orthonormal linear units on $\varphix$ and cross two axes (instantiated on natural images in \hyperref[sec:natural]{Section~\ref{sec:natural}}).
The member objective is either unsupervised (the OC objective, each unit a scalar projection trained to vanish in-distribution) or supervised (a cross-entropy classifier head). 
The scoring rule is either norm (RMS output magnitude) or disagreement (the spread of members about their mean). 
The norm rule uses one formula for both member types
\begin{equation}
    \label{eq:norm}
    S'_{\mathrm{N}}(x)
    =
    \sqrt{\frac{1}{K}\sum_{k=1}^{K}\frac{\lVert u_k(x)\rVert^2}{\dim u_k}}\,,
\end{equation}
with $u_k(x)=\mathbb{C}_k^\top\varphix$ (scalar, $\dim u_k{=}1$) for unsupervised units, reducing to $\lVert \mathbb{C}^\top\varphix\rVert/\sqrt{K}$, proportional to the OC norm (\autoref{eq:ocscore}), and $u_k(x)$ the classifier logits, one per class ($\dim u_k$ = the number of classes, so the RMS is over members and classes) for supervised units; it fills its whole column, including the \emph{supervised} $\times$ \emph{norm} hybrid cell (the corner that inverts).
Disagreement has no such single formula; classifier units use BALD on the softmax (the \emph{supervised} $\times$ \emph{disagreement} diagonal), and scalar units use output variance (the \emph{unsupervised} $\times$ \emph{disagreement} hybrid cell).

\smallskip\noindent\textbf{Diversity.} 
(i) \emph{Weight-space diversity} $\delta_{\mathrm{w}}$, one minus the mean absolute cosine between the flattened head weight vectors (high $=$ diverse), is the quantity the weight-orthonormality penalty directly maximizes.%
\footnote{%
    $\delta_{\mathrm{w}}=1-\operatorname{mean}_{i\neq j}|\cos(\hat w_i,\hat w_j)|$ measures weight decorrelation, not ensemble disagreement (we report function-space diversity separately for this reason); the penalty of \autoref{eq:ortho} uses $\cos^2$ rather than $|\cos|$, which are monotonically related.
} 
(ii) \emph{Function-space diversity (unbounded).}
The disagreement of the member predictions, quantified by the across-member prediction variance.
(iii) \emph{Function-space diversity (bounded).}
$\delta_{\mathrm{f}}=1-$mean pairwise linear CKA \citep{Kornblith::2019aa} on the member activations $z_k$ (\hyperref[sec:divobj]{Section~\ref{sec:divobj}}), the function-space analog of $\delta_{\mathrm{w}}$.
(iv) \emph{Per-member logit cosine (diagnostic).} The same $1-|\cos|$ on logit outputs, not reported as a diversity metric since it conflates weight- and function-space diversity.

\smallskip\noindent\textbf{Reporting and statistical testing.}
All aggregated results come from ten seeds controlling backbone and detector initialization and data order; compared models differ only in the variable under evaluation.
The default metric is ROC AUC, reported as a mean $\pm$ standard deviation and comparable within but not across experiments that change backbone depth or the spectral-normalization implementation.
Reported accuracies are test-set accuracy.
Every comparison is a paired difference across the shared seeds, labeled \emph{significant} when the paired $t$ clears $p<0.05$ for $n=10$ (two-tailed $|t|\ge2.26$, df$=9$), \emph{consistent} when all ten seeds agree in sign but the difference stays below that threshold, and \emph{within noise} otherwise.
In comparison tables, \textbf{bold} marks the significantly best value in each group; in ablation and descriptive tables it marks the value under discussion; \emph{italic} marks an inverted score ($<0.5$).
Every OOD score, namely BALD, Expected Pairwise KL divergence (EPKL), Variance-Gated Margin Uncertainty (VGMU), and their cov-LLE variants, comes from a single shared implementation, and every comparison and $|t|$ value from one significance-testing routine.

\section{Results}
\label{sec:results}
%
\subsection{The unifying $2\times2$ on natural images}
\label{sec:natural}
\begin{figure*}[t]
\centering
\includegraphics[width=\textwidth]{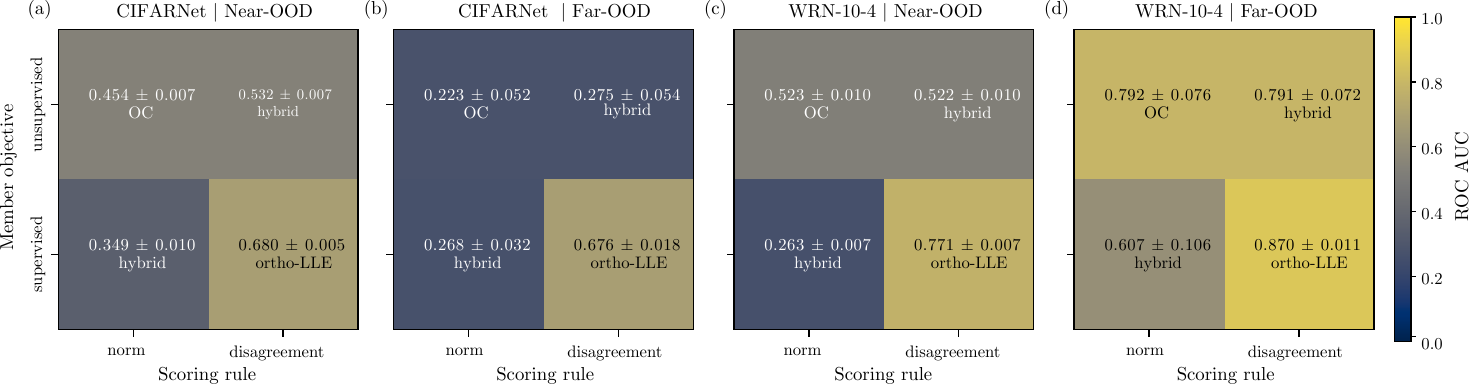}
\caption{%
    \textbf{The $\bm{2\times2}$ structure holds on natural images; its label-free corner is feature-quality gated.} Member objective (unsupervised/supervised) $\bm{\times}$ scoring rule (norm/disagreement) for the CIFARNet and WRN-10-4; near-OOD $\bm{=}$ held-out CIFAR 5--9, far-OOD $\bm{=}$ SVHN (ROC AUC). The supervised $\bm{\times}$ norm cell inverts on both, while the other three detect; the label-free OC norm (unsupervised $\bm{\times}$ norm) is near chance on the CIFARNet and recovers far-OOD only on the stronger WRN-10-4.
}
\label{fig:natural_image_2x2}
\end{figure*}

We establish the $2\times2$ structure directly on natural images (ID $=$ CIFAR-10 classes 0--4, near-OOD $=$ held-out classes 5--9, far-OOD $=$ SVHN), first on the CIFARNet and then on a WRN-10-4, freezing each backbone and running identical detectors and the $2\times2$ (\autoref{fig:natural_image_2x2}).
On both backbones, orthonormality restores the collapsed diversity (weight-space diversity, vanilla$\to$ortho $0.037\!\to\!0.981$ on the CIFARNet and $0.038\!\to\!0.970$ on the WRN-10-4, \autoref{tab:wrn_recovery}). Supervised disagreement (BALD) is the score that detects: three of the four cells are informative, and only \emph{supervised} $\times$ \emph{norm} inverts (\emph{unsupervised} $\times$ \emph{norm} $\approx$ \emph{unsupervised} $\times$ \emph{disagreement}).
However, the label-free OC norm fails on the CIFARNet, with near-OOD at chance ($0.454$) and far-OOD SVHN strongly \emph{inverted} ($0.223$), the likelihood paradox \citep{Nalisnick::2019aa}, in which a learned magnitude ranks SVHN as more typical than CIFAR.
The fault is the features, not the detector: replacing the CIFARNet (accuracy: $0.821$) with a WRN-10-4 (accuracy: $0.935$) recovers far-OOD OC from an inverted $0.223$ to $0.792$ ($t\approx19$, \autoref{tab:wrn_recovery}), with the ortho-LLE best on far-OOD.
Near-OOD OC stays at chance under norm (magnitude) scoring, which the direction score repairs (\hyperref[sec:direction]{Section~\ref{sec:direction}}).
\begin{table*}[h]
\centering
\caption{%
    \textbf{A stronger backbone recovers the label-free OC score.} CIFARNet vs.\ WRN-10-4 (ROC AUC): far-OOD OC rises from an inverted $\bm{0.223}$ to $\bm{0.792}$ and orthonormality restores weight-space diversity ($\bm{\delta_{\mathrm{w}}}$) on both, while near-OOD OC remains at chance.\textsuperscript{a}
    }
\label{tab:wrn_recovery}
\footnotesize
\begin{tabular*}{\textwidth}{@{\extracolsep{\fill}}l*{4}{c}@{}}
\toprule
 & \multicolumn{2}{c}{\textbf{CIFARNet}} & \multicolumn{2}{c}{\textbf{WRN-10-4}} \\
\cmidrule(lr){2-3}\cmidrule(lr){4-5}
\textbf{Detector} & near & far & near & far \\
\midrule
in-distribution accuracy & \multicolumn{2}{c}{$0.821\pm0.012$} & \multicolumn{2}{c}{$0.935\pm0.002$} \\
\midrule
OC        & $0.454\pm0.014$ & $\mathit{0.223\pm0.047}$ & $0.523\pm0.010$ & $0.792\pm0.076$ \\
van-LLE   & $0.646\pm0.009$ & $\bm{0.697\pm0.026}$ & $0.746\pm0.006$ & $0.845\pm0.026$ \\
ortho-LLE & $\bm{0.656\pm0.011}$ & $0.685\pm0.025$ & $\bm{0.771\pm0.007}$ & $\bm{0.869\pm0.011}$ \\
\midrule
weight-space diversity ($\delta_{\mathrm{w}}$: vanilla $\to$ ortho) & \multicolumn{2}{c}{$0.037\pm0.006 \to 0.981\pm0.003$} & \multicolumn{2}{c}{$0.038\pm0.002 \to 0.970\pm0.001$} \\
\bottomrule
\end{tabular*}
\vspace{2pt}
\noindent\parbox{\textwidth}{%
    \footnotesize\textsuperscript{a} 
    CIFAR-10: 0--4 in-distribution / 5--9 near-OOD / SVHN far-OOD.
    \textbf{Bold} $=$ significantly-best detector per column (paired $t$, $p<0.05$, $n=10$); \emph{italic} $=$ inverted ($<0.5$).
}
\end{table*}

\subsection{Diversity in function space: An objective that outperforms the weight-space penalty}
\label{sec:divobj}
Shared gradients can pull the members of a last-layer ensemble toward one function, so their disagreement collapses, and with it the signal ensemble uncertainty depends on.
The weight-orthonormality of OC repairs this collapse in weight space, but the correction reaches the predictions of the members only partially (\hyperref[app:divobj]{SI~\ref{app:divobj}}).
This section answers the open question of whether last-layer diversity can close that gap, and what mitigates the collapse \citep{Kirsch::2025aa}.

\begin{figure}[h]
\centering
\includegraphics[width=\columnwidth]{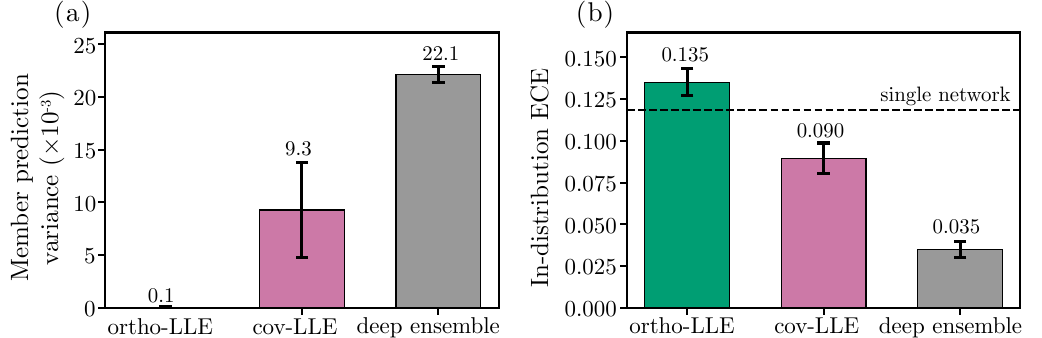}
\caption{%
    \textbf{Cov-LLE is between the weight-orthonormal LLE and a deep ensemble on both prediction variance and calibration.} Member prediction variance (a) and in-distribution ECE (b) for ortho-LLE, cov-LLE, and a $\bm{K{=}10}$ deep ensemble on the CIFARNet; dashed line: ECE for a single network.
    }
\label{fig:collapse_spectrum}
\end{figure}

\smallskip\noindent\textbf{A function-space objective.} 
Give each member a small embedding $z_k(x)\in\mathbb{R}^{D}$ ($\varphi\to z_k\to$ logits) and decorrelate the \emph{activations} of the members rather than their weights, with a Barlow-Twins / VICReg-style \citep{Zbontar::2021aa,Bardes::2022aa} covariance penalty on the centered, concatenated embeddings $Z\in\mathbb{R}^{N\times KD}$, where
\begin{equation} 
    \label{eq:cov} 
    \Omega_{\mathrm{cov}}(Z) 
    = 
    \tfrac{1}{KD}\big\lVert \tfrac{1}{N}Z^\top Z - I\big\rVert_F^2 
\end{equation} 
is added to cross-entropy with weight $\lambda_\mathrm{cov}$, defining the covariance LLE (cov-LLE).
At matched $K$ a deep ensemble has orders of magnitude more in-distribution prediction variance than the shared-backbone ortho-LLE and can out-calibrate it (\autoref{fig:collapse_spectrum}).
This is the function-space diversity the ortho-LLE fails to attain.
Targeting that diversity directly in function space, cov-LLE recovers much of it at $1\times$ backbone cost and significantly outperforms the weight-space ortho-LLE.

\smallskip\noindent\textbf{On strong CIFAR backbones: A robust far-OOD gain.} 
The decisive test compares the cov-LLE against the $K{=}32$ ortho-LLE (\hyperref[app:openood]{SI~\ref{app:openood}}), on the same frozen features. 
\autoref{tab:covlle_vs_ortho} reports both on a frozen ResNet-18 (CIFAR-10, accuracy: $0.940$) and a frozen WRN-28-10 (accuracy: $0.950$). 
\begin{table*}[h]
\centering
\caption{%
    \textbf{Function-space decorrelation (cov-LLE) significantly beats the weight-orthonormal LLE on far-OOD on both frozen backbones, at no accuracy cost.} ROC AUC (BALD / EPKL / VGMU) and in-distribution ECE for cov-LLE ($\bm{\lambda_\mathrm{cov}{=}0.5}$), ortho-LLE, and a penalty-free two-layer LLE.\textsuperscript{a}
    }
\label{tab:covlle_vs_ortho}
\footnotesize
\begin{tabular*}{\textwidth}{@{\extracolsep{\fill}}ll*{6}{c}c@{}}
\toprule
& & \multicolumn{2}{c}{\textbf{BALD}} & \multicolumn{2}{c}{\textbf{EPKL}} & \multicolumn{2}{c}{\textbf{VGMU}} & \multirow{2}{*}{\textbf{ECE}} \\
\cmidrule(lr){3-4}\cmidrule(lr){5-6}\cmidrule(lr){7-8}
\textbf{Backbone} & \textbf{Detector} & near & far & near & far & near & far & \\
\midrule
\multirow{3}{*}{ResNet-18 (0.940)}
 & two-layer LLE          & $0.725\pm0.004$ & $0.758\pm0.010$ & $0.714\pm0.005$ & $0.765\pm0.009$ & $0.692\pm0.005$ & $0.719\pm0.012$ & $0.047\pm0.001$ \\
 & ortho-LLE              & $0.843\pm0.004$ & $0.892\pm0.006$ & $0.842\pm0.004$ & $0.893\pm0.005$ & $0.833\pm0.005$ & $0.885\pm0.008$ & $0.044\pm0.001$ \\
 & cov-LLE ($\lambda{=}0.5$) & $\bm{0.858\pm0.002}$ & $\bm{0.912\pm0.003}$ & $\bm{0.858\pm0.001}$ & $\bm{0.911\pm0.003}$ & $\bm{0.852\pm0.003}$ & $\bm{0.908\pm0.003}$ & $\bm{0.008\pm0.002}$ \\
\midrule
\multirow{3}{*}{WRN-28-10 (0.950)}
 & two-layer LLE          & $0.870\pm0.002$ & $0.899\pm0.007$ & $0.870\pm0.002$ & $0.900\pm0.006$ & $0.851\pm0.002$ & $0.880\pm0.010$ & $0.032\pm0.001$ \\
 & ortho-LLE              & $0.878\pm0.001$ & $0.922\pm0.004$ & $0.878\pm0.001$ & $0.922\pm0.004$ & $0.873\pm0.000$ & $0.913\pm0.002$ & $\bm{0.029\pm0.001}$ \\
 & cov-LLE ($\lambda{=}0.5$) & $0.881\pm0.005$ & $\bm{0.946\pm0.007}$ & $0.881\pm0.005$ & $\bm{0.947\pm0.007}$ & $0.872\pm0.006$ & $\bm{0.920\pm0.009}$ & $0.034\pm0.004$ \\
\bottomrule
\end{tabular*}
\vspace{2pt}
\noindent\parbox{\textwidth}{%
    \footnotesize\textsuperscript{a}
    ResNet-18 ($K=32$): near $=$ CIFAR-100 / TinyImageNet, far $=$ SVHN / MNIST / DTD; 
    WRN-28-10 ($K=32$): near $=$ CIFAR-100, far $=$ SVHN. 
    ECE is in-distribution. 
    \textbf{Bold} $=$ within each backbone, the significantly-best detector per column (paired $t$, $p<0.05$, $n=10$).
}
\end{table*}
The cov-LLE (at $\lambda_\mathrm{cov}{=}0.5$) significantly outperforms the ortho-LLE on far-OOD on both backbones (ResNet-18 $0.892\!\to\!0.912$, $t\approx8$; WRN-28-10 $0.922\!\to\!0.946$, $t\approx11$) at unchanged accuracy. 
The near-OOD and calibration gains are real but backbone-dependent, largest where the weight-space penalty leaves the most room to improve.
On the weaker ResNet-18 cov-LLE also significantly outperforms on near-OOD ($0.843\!\to\!0.858$, $t\approx10$) and improves calibration (ECE $0.044\!\to\!0.008$, $t\approx35$); on the stronger WRN-28-10 near-OOD is a tie ($0.878\!\to\!0.881$) and calibration marginally worse ($0.029\!\to\!0.034$). 
A matched two-layer head without the penalty does not achieve any of these gains (it collapses to $0.725/0.758$ on ResNet-18), so the gain is the covariance objective, not the extra layer.

\begin{figure*}[h]
\centering
\includegraphics[width=\textwidth]{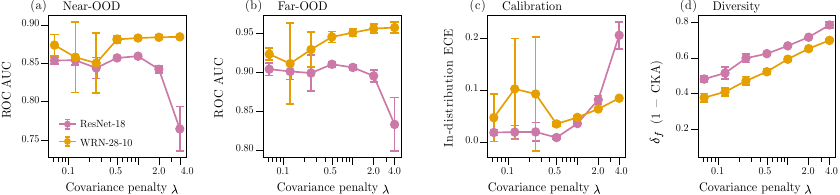}
\caption{%
    \textbf{Detection peaks at an interior $\bm{\lambda_\mathrm{cov}{\approx}0.5}$ and collapses as the penalty over-decorrelates, while function-space diversity keeps rising.} Covariance-penalty strength $\bm{\lambda_\mathrm{cov}}$ on two frozen backbones, ResNet-18 and WRN-28-10, showing near-OOD, far-OOD, in-distribution ECE, and diversity $\bm{\delta_{\mathrm{f}}}$. ID $=$ CIFAR-10, OpenOOD protocol (\hyperref[app:openood]{SI~\ref{app:openood}}).
    }
\label{fig:covlle_lambda_sweep}
\end{figure*}

\emph{Why act on activations rather than weights?} 
The penalty works because it decorrelates a quantity the weight-space penalty cannot affect. 
This is clearest on MNIST 0--4 with this two-layer head (frozen $\varphi$, $K{=}32$, $\lambda_\mathrm{cov}{=}0.5$, 10 seeds; \hyperref[app:divobj]{SI~\ref{app:divobj}}, \autoref{tab:covlle_mnist}), where weight-orthonormality leaves members functionally identical (function-space diversity $\delta_{\mathrm{f}}$ only $0.015$, indistinguishable from the $0.015$ of a plain ensemble) because it acts on an embedding map the classifier layer above negates. 
The covariance penalty decorrelates ($\delta_{\mathrm{f}}$ $0.015\!\to\!0.520$, member prediction variance ${\sim}6\times$) and raises near-OOD ROC AUC $0.888\!\to\!0.944$ at no accuracy cost (full frozen/joint breakdown in \hyperref[app:divobj]{SI~\ref{app:divobj}}, \autoref{tab:covlle_mnist}).
The gain is not MNIST-specific, since on the strong ResNet-18 the penalty lifts $\delta_{\mathrm{f}}$ from $0.077$ (plain two-layer head) to $0.623$ at $\lambda_\mathrm{cov}{=}0.5$ (\autoref{fig:covlle_lambda_sweep}).

\emph{There is an interior optimum.} 
As \autoref{fig:covlle_lambda_sweep} varies $\lambda_\mathrm{cov}$, detection rises to a plateau and then collapses as the penalty over-decorrelates (ResNet-18 falls to $0.765/0.833$ at $\lambda_\mathrm{cov}{=}4$, ECE increasing up to $0.206$). 
At the detection-optimal $\lambda_\mathrm{cov}{\approx}0.5$ the ResNet-18 ECE is in fact better than that of the ortho-LLE, not worse; a ``calibration cost'' only appears at over-penalizing values of $\lambda_\mathrm{cov}$. 

\smallskip\noindent\textbf{It recovers the benefits of a deep ensemble at $\bm{1\times}$ cost.} 
What a deep ensemble adds over the shared-backbone LLE is function-space diversity, and that improves calibration where gains exist.
The covariance objective recovers a large part of that without $K\times$ independent networks. 
At matched $K{=}10$ its in-distribution prediction variance resides between the ortho-LLE and a deep ensemble (CIFAR $0.05\!\to\!9.3\!\to\!22.1$, $\times10^{-3}$, for ortho$\to$cov$\to$deep), as does its calibration (ECE $0.135\!\to\!0.090\!\to\!0.035$; cov-LLE is significantly better-calibrated than ortho, $t\approx10$, \autoref{fig:collapse_spectrum}), a single-forward-pass head approaching a full deep ensemble on the signature benefit (\autoref{tab:deep_ensemble_comparison}). 
On these weak backbones the recovered diversity does not raise OOD ROC AUC; that gain requires strong feature representations.
\begin{table*}[h]
\centering
\caption{%
    \textbf{At $\bm{1\times}$ backbone cost, cov-LLE recovers much of the function-space diversity of a deep ensemble (and its calibration on the harder CIFAR task), is between the ortho-LLE and the $\bm{K}$-network deep ensemble.} Prediction variance, ROC AUC (BALD / EPKL / VGMU), and in-distribution ECE at matched $\bm{K{=}10}$ on MNIST 0--4 and CIFAR 0--4.\textsuperscript{a}
    }
\label{tab:deep_ensemble_comparison}
\footnotesize
\begin{tabular*}{\textwidth}{@{\extracolsep{\fill}}lc*{6}{c}c@{}}
\toprule
& \textbf{Prediction variance} & \multicolumn{2}{c}{\textbf{BALD / MSP}} & \multicolumn{2}{c}{\textbf{EPKL}} & \multicolumn{2}{c}{\textbf{VGMU}} & \multirow{2}{*}{\textbf{ECE}} \\
\cmidrule(lr){3-4}\cmidrule(lr){5-6}\cmidrule(lr){7-8}
\textbf{Method} & ($\times10^{-3}$) & near & far & near & far & near & far & \\
\midrule
\multicolumn{9}{l}{\emph{MNIST 0--4 (MNISTNet; accuracy 0.998)}} \\
single network    & --- & $0.908\pm0.013$ & $0.986\pm0.004$ & --- & --- & --- & --- & $0.002\pm0.000$ \\
ortho-LLE     & $0.008\pm0.010$ & $0.931\pm0.006$ & $0.992\pm0.001$ & $0.932\pm0.006$ & $0.993\pm0.001$ & $0.925\pm0.006$ & $0.990\pm0.002$ & $\bm{0.001\pm0.000}$ \\
cov-LLE       & $\bm{0.900\pm0.220}$ & $0.931\pm0.015$ & $0.978\pm0.006$ & $0.933\pm0.014$ & $0.979\pm0.006$ & $0.928\pm0.016$ & $0.977\pm0.007$ & $0.009\pm0.001$ \\
deep ensemble & $0.441\pm0.135$ & $0.937\pm0.010$ & $0.992\pm0.002$ & $0.937\pm0.011$ & $0.991\pm0.003$ & $0.935\pm0.008$ & $\bm{0.993\pm0.001}$ & $0.004\pm0.001$ \\
\midrule
\multicolumn{9}{l}{\emph{CIFAR 0--4 (CIFARNet; accuracy 0.810)}} \\
single network    & --- & $0.623\pm0.008$ & $\bm{0.723\pm0.033}$ & --- & --- & --- & --- & $0.119\pm0.009$ \\
ortho-LLE     & $0.054\pm0.019$ & $0.644\pm0.009$ & $0.679\pm0.021$ & $0.648\pm0.009$ & $\bm{0.684\pm0.020}$ & $0.634\pm0.009$ & $0.728\pm0.026$ & $0.135\pm0.008$ \\
cov-LLE       & $9.270\pm4.535$ & $0.591\pm0.009$ & $0.605\pm0.041$ & $0.586\pm0.010$ & $0.591\pm0.043$ & $0.600\pm0.012$ & $0.640\pm0.041$ & $0.090\pm0.009$ \\
deep ensemble & $\bm{22.140\pm0.790}$ & $\bm{0.666\pm0.007}$ & $0.672\pm0.012$ & $\bm{0.663\pm0.007}$ & $0.645\pm0.014$ & $\bm{0.664\pm0.007}$ & $0.723\pm0.013$ & $\bm{0.035\pm0.005}$ \\
\bottomrule
\end{tabular*}
\vspace{2pt}
\noindent\parbox{\textwidth}{%
    \footnotesize\textsuperscript{a}
    Single network evaluated by maximum softmax probability (MSP).
    \textbf{Bold} $=$ within each dataset block, the significantly-best method per column (paired $t$, $p<0.05$, $n=10$).
}
\end{table*}

The pattern generalizes beyond these two backbones, to the OpenOOD suite, four RobustBench checkpoints, and a pretrained CIFAR-100 backbone (\hyperref[sec:featquality]{Section~\ref{sec:featquality}}, \hyperref[sec:baselines]{Section~\ref{sec:baselines}}, \hyperref[app:openood]{SI~\ref{app:openood}}, \hyperref[app:robustbench]{SI~\ref{app:robustbench}}), and where it does not the cause is feature quality, not the mechanism (a weak CIFARNet, where the cov-LLE scores below the ortho-LLE, \autoref{tab:deep_ensemble_comparison}).
\emph{Diversity is the mechanism, it must be enforced in function space, and doing so provides the strongest of the OC and last-layer scores we measure, while inexpensively recovering the diversity and calibration of a deep ensemble.}

\subsection{Feature quality, not conditioning, controls the label-free detector}
\label{sec:featquality}
\hyperref[sec:natural]{Section~\ref{sec:natural}} attributes the far-OOD competence of OC to feature quality, but from a single backbone pair. 
The natural-image experiments conflate feature quality with Lipschitz conditioning. 
We separate the two with post-hoc probes on standard backbones (\hyperref[app:openood]{SI~\ref{app:openood}}, \hyperref[app:robustbench]{SI~\ref{app:robustbench}}) and a controlled spectral-normalization ablation (\hyperref[app:snablation]{SI~\ref{app:snablation}}), and all three agree:
(i) Under the OpenOOD CIFAR-10 protocol (every detector on one frozen ResNet-18), the ortho-LLE significantly outperforms the van-LLE on every scoring rule (BALD near $0.812\to0.843$, far $0.858\to0.892$). 
However, the label-free OC norm (magnitude) is weak on this vanilla, non-bi-Lipschitz ResNet-18 ($0.680/0.790$ near/far), consistent with the feature-quality account (the full baseline ranking is in \hyperref[sec:baselines]{Section~\ref{sec:baselines}}).
(ii) Holding architecture fixed (WRN-28-10) and varying only training (standard vs.\ adversarially robust) via RobustBench \citep{Croce::2021aa}, more-robust backbones are significantly worse (\autoref{tab:robustbench_inset}). 
The label-free OC norm collapses on far-OOD ($0.839$ standard to $0.509$--$0.622$ robust, near chance), while the supervised ensembles are more resilient (ortho-LLE $0.922\to0.836$--$0.875$, cov-LLE $0.946\to0.893$--$0.919$); detection tracks clean accuracy ($0.950\to0.874$--$0.897$), not adversarial robustness.
(iii) A fixed-architecture \{plain, $+$SN, $+$SN$+$residual\} ablation on the CIFARNet (\autoref{tab:cifar_spectralnorm_ablation}) confirms the asymmetry.
Spectral normalization plus a residual lifts the supervised ortho-LLE on far-OOD (BALD $0.676\to0.737$) but leaves the label-free OC norm inverted (far $0.305\to0.202$, still below chance), so smoothness helps the supervised ensemble, while the label-free norm is neither repaired nor required by it.
Therefore, feature quality, not smoothness, controls the label-free score, and adversarial robustness trades away the feature quality the detector requires.
\begin{table}[h]
\centering
\caption{%
    \textbf{Adversarial robustness hurts far-OOD detection: The standard checkpoint significantly outperforms all three $\ell_\infty$-robust WRN-28-10s.} Far-OOD (SVHN) ROC AUC for OC, ortho-LLE, and cov-LLE.\textsuperscript{a}
    }
\label{tab:robustbench_inset}
\footnotesize
\setlength{\tabcolsep}{3pt}
\begin{tabular*}{\columnwidth}{@{\extracolsep{\fill}}lcccc@{}}
\toprule
\textbf{WRN-28-10} & \textbf{Clean accuracy} & \textbf{OC} & \textbf{ortho-LLE} & \textbf{cov-LLE} \\
\midrule
Standard & $\bm{0.950}$ & $\bm{0.839\pm0.070}$ & $\bm{0.922\pm0.004}$ & $\bm{0.946\pm0.007}$ \\
Carmon   & $0.897$ & $0.622\pm0.047$ & $0.872\pm0.008$ & $0.908\pm0.004$ \\
Gowal    & $0.894$ & $0.509\pm0.025$ & $0.875\pm0.006$ & $0.919\pm0.003$ \\
Rebuffi  & $0.874$ & $0.577\pm0.034$ & $0.836\pm0.008$ & $0.893\pm0.005$ \\
\bottomrule
\end{tabular*}
\vspace{2pt}
\noindent\parbox{\columnwidth}{%
    \footnotesize\textsuperscript{a}
    Full table in \hyperref[app:robustbench]{SI~\ref{app:robustbench}}.
}
\end{table}

\subsection{A direction score corrects the label-free near-OOD failure}
\label{sec:direction}
\begin{figure}[t]
\centering
\includegraphics[width=\columnwidth]{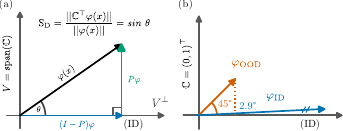}
\caption{%
    The direction score measures the angle a feature makes with the in-distribution subspace, not its length.
    (a) The sine identity. With the certificate span $\bf{V}\bf{=}\operatorname{\bf{span}}(\mathbb{C})$ and the ID subspace $\bf{V}^\perp$ its orthogonal complement, the OC norm $\mathrm{\bf{S_N}}=\boldsymbol{\lVert}\mathbb{C}^\top\boldsymbol{\varphix}\boldsymbol{\rVert}$ is the edge opposite the angle $\boldsymbol{\theta}$ that $\boldsymbol{\varphix}$ makes with $\bf{V}^\perp$, and $\mathrm{\bf{S_D}}=\mathrm{\bf{S_N}}\,/\,\boldsymbol{\lVert}\boldsymbol{\varphix}\boldsymbol{\rVert}=\boldsymbol{\sin\theta}$.
    (b) A two-dimensional example. An ID point $\boldsymbol{\varphi}_{\mathrm{\bf{ID}}}\bf{=(10,0.5)}$ nearly aligned with the ID axis and an OOD point $\boldsymbol{\varphi}_{\mathrm{\bf{OOD}}}\bf{=(0.3,0.3)}$ pointing $\bm{45^\circ}$ off it. Magnitude inverts the ranking, placing ID above OOD ($\mathrm{\bf{S_N}}\bf{=0.5>0.3}$), while direction restores the correct order ($\mathrm{\bf{S_D}}\bf{\approx0.05<0.71}$).
}
\label{fig:direction_sine}
\end{figure}

The label-free OC score is the certificate norm $\mathrm{S_N}(x)=\lVert \mathbb{C}^\top\varphix\rVert$ (the OC norm of \hyperref[sec:bg]{Section~\ref{sec:bg}}).
An input is flagged out-of-distribution when its feature has a large component along the certificate directions.
However, a component can be large for two unrelated reasons: (i) the feature \emph{points away} from the in-distribution subspace (the anomaly we want), or (ii) the feature is simply a \emph{long vector}.
The $\mathrm{S_N}$ (a magnitude score) cannot separate them, ranking large features above small-but-anomalous ones.
The solution is to score \emph{direction}, not magnitude.
The scale-invariant ${\mathrm{S_D}}(x)=\lVert \mathbb{C}^\top\varphix\rVert \,/\,\lVert\varphix\rVert=\sin\theta$ keeps only the angle $\theta$ between $\varphix$ and the in-distribution subspace (\autoref{fig:direction_sine}(a)).

For orthonormal certificates ($\mathbb{C}^\top\mathbb{C}=I_K$) the projector $P=\mathbb{C}\mathbb{C}^\top$ onto $V=\operatorname{span}(\mathbb{C})$ makes $\lVert\mathbb{C}^\top\varphix\rVert=\lVert P\varphix\rVert$, the edge opposite the angle $\theta$ that $\varphix$ makes with the in-distribution subspace $V^\perp$, so dividing by the hypotenuse $\lVert\varphix\rVert$ results in $\mathrm{S_D}=\sin\theta$.
Panel~(b) shows why this matters.
With in-distribution data along ID (axis-1), a single certificate $\mathbb{C}=(0,1)^\top$ reads off the second coordinate, and for a large ID point $\varphix=(10,0.5)$ and a small OOD point $\varphix=(0.3,0.3)$ the magnitude score inverts, ranking ID above OOD ($\mathrm{S_N}=0.5>0.3$).
Meanwhile, the direction score restores the order ($\mathrm{S_D}=\sin2.9^\circ\approx0.05<\sin45^\circ\approx0.71$): dividing by $\lVert\varphix\rVert$ adjusts for the near-alignment of the ID point with the in-distribution axis.
Across four frozen backbones (\autoref{tab:direction_score}; certificates re-fit separately from \autoref{tab:wrn_recovery}, so magnitude baselines differ slightly; full analysis in \hyperref[app:direction]{SI~\ref{app:direction}}), near- and far-OOD split cleanly.

\smallskip\noindent\textbf{Near-OOD.} Direction adds $+0.160$ to $+0.178$ ROC AUC on every backbone (all $t\ge33$); the failure was a scoring artifact, not a capacity limit.
The certificates carried the signal and the magnitude score discarded it.

\smallskip\noindent\textbf{Far-OOD.} Direction removes the SVHN inversion wherever magnitude misleads (CIFARNet $0.218\!\to\!0.511$, ResNet-18 $0.677\!\to\!0.872$, WRN-28-10 $0.806\!\to\!0.922$; all significant). The exception is WRN-10-4, where its recovery detects SVHN through certificate \emph{magnitude}, so dividing that out inverts the score. 
The two are thus complementary, and a label-free max-fusion (\hyperref[app:direction]{SI~\ref{app:direction}}) never inverts and recovers most of both. 
\begin{table*}[h]
\centering
\caption{%
    \textbf{The direction of OC certificates is more discriminative than their magnitude on near-OOD for every backbone, and on far-OOD for the stronger ones; the two scores are complementary.} ROC AUC for magnitude, direction, and their max-fusion using OC certificates ($\bm{K{=}32}$, $\bm{\lambda_\mathrm{oc}{=}5}$).\textsuperscript{a}
    }
\label{tab:direction_score}
\footnotesize
\begin{tabular*}{\textwidth}{@{\extracolsep{\fill}}l*{3}{c}*{3}{c}@{}}
\toprule
& \multicolumn{3}{c}{\textbf{Near-OOD}} & \multicolumn{3}{c}{\textbf{Far-OOD}} \\
\cmidrule(lr){2-4}\cmidrule(lr){5-7}
\textbf{Backbone} 
& magnitude
& direction 
& max-fusion 
& magnitude
& direction 
& max-fusion 
\\
\midrule
CIFARNet (0.821)  & $\mathit{0.446\pm0.010}$ & $\bm{0.624\pm0.012}$ & $0.564\pm0.012$ & $\mathit{0.218\pm0.052}$ & $\bm{0.511\pm0.057}$ & $\mathit{0.389\pm0.063}$ \\
WRN-10-4 (0.935)  & $0.529\pm0.008$ & $\bm{0.689\pm0.006}$ & $0.642\pm0.007$ & $\bm{0.794\pm0.040}$ & $\mathit{0.300\pm0.069}$ & $0.710\pm0.050$ \\
ResNet-18 (0.940) & $0.605\pm0.032$ & $\bm{0.778\pm0.022}$ & $0.759\pm0.024$ & $0.677\pm0.047$ & $\bm{0.872\pm0.019}$ & $0.855\pm0.022$ \\
WRN-28-10 (0.950) & $0.581\pm0.040$ & $\bm{0.746\pm0.038}$ & $0.730\pm0.039$ & $0.806\pm0.046$ & $\bm{0.922\pm0.032}$ & $0.912\pm0.034$ \\
\bottomrule
\end{tabular*}
\vspace{2pt}
\noindent\parbox{\textwidth}{%
    \footnotesize\textsuperscript{a}
     Accuracy in parentheses. 
     \textbf{Bold} $=$ per backbone, near and far, the significantly-best of the three scores (paired $t$, $p<0.05$, $n=10$); \emph{italic} $=$ inverted ($<0.5$).
}
\end{table*}

\subsection{Comparison to standard OOD baselines}
\label{sec:baselines}
So that these scores are not evaluated in isolation, \autoref{tab:ood_baselines} places OC and the last-layer ensembles against standard post-hoc detectors on a single frozen ResNet-18 (CIFAR-10, accuracy: $0.940$; OpenOOD protocol \citep{Yang::2022aa}, \autoref{tab:ood_baselines}).
\begin{table*}[h]
\centering
\caption{%
    \textbf{cov-LLE is the best last-layer score and competitive with the post-hoc baselines at $\bm{O(KC)}$ inference cost; KNN leads on raw scores.} OC and the last-layer ensembles vs.\ standard post-hoc detectors on one frozen ResNet-18 (OpenOOD protocol).\textsuperscript{a}
}
\label{tab:ood_baselines}
\footnotesize
\setlength{\tabcolsep}{4pt}
\begin{tabular*}{\textwidth}{@{\extracolsep{\fill}}ll cccc ll@{}}
\toprule
& & \multicolumn{2}{c}{\textbf{ROC AUC}$^\uparrow$} & \multicolumn{2}{c}{\textbf{FPR@95}$^\downarrow$} & \multicolumn{2}{c}{\textbf{Complexity}} \\
\cmidrule(lr){3-4}\cmidrule(lr){5-6}\cmidrule(lr){7-8}
\textbf{Detector} & \textbf{Score} & near & far & near & far & space & time \\
\midrule
\multicolumn{8}{@{}l}{\emph{Baselines (deterministic)}} \\
MSP \cite{Hendrycks::2017aa}   & & $0.859$ & $0.909$ & $0.531$ & $0.282$ & $O(1)$ & $O(C)$ \\
MaxLogit                       & & $0.852$ & $0.914$ & $0.675$ & $0.364$ & $O(1)$ & $O(C)$ \\
Energy \cite{Liu::2020ab}      & & $0.853$ & $0.915$ & $0.676$ & $0.364$ & $O(1)$ & $O(C)$ \\
Mahalanobis \cite{Lee::2018ac} & & $0.820$ & $0.920$ & $0.580$ & $0.252$ & $O(d^2{+}Cd)$ & $O(d^2{+}Cd)$ \\
KNN \cite{Sun::2022aa}         & & $\bm{0.887}$ & $\bm{0.934}$ & $\bm{0.430}$ & $\bm{0.218}$ & $O(Nd)$ & $O(Nd)$ \\
\midrule
\multicolumn{8}{@{}l}{\emph{OC and last-layer ensembles}} \\
\multirow{2}{*}{OC} & magnitude & $0.680\pm0.008$ & $0.790\pm0.014$ & $0.819\pm0.011$ & $0.604\pm0.042$ & $O(Cd)$ & $O(Cd)$ \\
 & direction & $0.807\pm0.004$ & $0.906\pm0.005$ & $0.681\pm0.011$ & $0.358\pm0.020$ & $O(Cd)$ & $O(Cd)$ \\
\midrule
\multirow{3}{*}{van-LLE} & BALD & $0.812\pm0.004$ & $0.858\pm0.004$ & $0.900\pm0.003$ & $0.900\pm0.003$ & $O(KdC)$ & $O(KC)$ \\
 & EPKL & $0.817\pm0.004$ & $0.870\pm0.003$ & $1.000$ & $1.000$ & $O(KdC)$ & $O(K^2C)$ \\
 & VGMU & $0.785\pm0.006$ & $0.831\pm0.007$ & $1.000$ & $1.000$ & $O(KdC)$ & $O(C)$ \\
\midrule
\multirow{3}{*}{ortho-LLE} & BALD & $0.843\pm0.004$ & $0.892\pm0.006$ & $0.769\pm0.076$ & $0.343\pm0.031$ & $O(KdC)$ & $O(KC)$ \\
 & EPKL & $0.842\pm0.004$ & $0.893\pm0.005$ & $0.980\pm0.063$ & $0.345\pm0.029$ & $O(KdC)$ & $O(K^2C)$ \\
 & VGMU & $0.833\pm0.005$ & $0.885\pm0.008$ & $1.000$ & $0.622\pm0.193$ & $O(KdC)$ & $O(C)$ \\
\midrule
\multirow{3}{*}{cov-LLE} & BALD & $\bm{0.858\pm0.002}$ & $\bm{0.912\pm0.003}$ & $0.663\pm0.023$ & $0.288\pm0.014$ & $O(KdC)$ & $O(KC)$ \\
 & EPKL & $\bm{0.858\pm0.001}$ & $\bm{0.911\pm0.003}$ & $\bm{0.657\pm0.023}$ & $\bm{0.283\pm0.015}$ & $O(KdC)$ & $O(K^2C)$ \\
 & VGMU & $0.852\pm0.003$ & $0.908\pm0.003$ & $0.716\pm0.022$ & $0.316\pm0.021$ & $O(KdC)$ & $O(C)$ \\
\bottomrule
\end{tabular*}
\vspace{2pt}
\noindent\parbox{\textwidth}{%
    \footnotesize\textsuperscript{a}
    ID $=$ CIFAR-10 (accuracy 0.940); near-OOD $=$ \{CIFAR-100, TinyImageNet\}, far-OOD $=$ \{MNIST, SVHN, DTD\}; ROC AUC / FPR@95 averaged over the datasets in each group (10 head seeds; baselines deterministic).
    Space and time are the extra state and per-query cost beyond the shared forward pass ($N$ train points, $C$ classes, $d$ feature dimension, $K$ heads); only KNN scales with $N$.
    The OC \emph{direction} score (\autoref{sec:direction}) and cov-LLE (\autoref{sec:divobj}) are our proposed scores.
    \textbf{Bold} $=$ best per column, bolded independently within each block: numeric best for the deterministic baselines; for the OC and LLE block, the significantly-best (paired $t$, $p<0.05$, $n=10$): cov-BALD/EPKL on ROC AUC (indistinguishable from each other, \autoref{sec:baselines}) and cov-EPKL on FPR@95.
}
\end{table*}

On raw ROC AUC the last-layer ensembles rank in the middle of the baselines. 
The ortho-LLE ($0.843/0.892$ near/far) is competitive with the logit-space MSP \citep{Hendrycks::2017aa} and energy \citep{Liu::2020ab} but does not outperform them, and all rank below the feature-space KNN ($0.887/0.934$, \citep{Sun::2022aa}).
Our two proposed scores are the strongest of the OC and last-layer results. 
The function-space cov-LLE (\hyperref[sec:divobj]{Section~\ref{sec:divobj}}) reaches $0.858/0.912$ near/far, best in that group, and the direction score (\hyperref[sec:direction]{Section~\ref{sec:direction}}) raises OC above its magnitude score.
We do not claim state-of-the-art (KNN leads on raw ROC AUC); the contribution is the \emph{mechanism} (diversity in function space) and the \emph{diagnosis} (the failures of OC are scoring artifacts). 
The last-layer ensemble also fills a niche a pure post-hoc detector does not. 
Unlike KNN, which stores and queries the full training feature bank at test time, it delivers a calibrated predictive distribution and its epistemic (BALD) decomposition in a single frozen-backbone pass with no stored reference set, and falls back to a label-free score (OC with the direction score) when labels are unavailable.

\subsection{The mechanism transfers across modality and scale}
\label{sec:generality}
So far, we have only presented results for computer vision.
Since the last-layer ensemble needs only frozen features $\varphi$, it should transfer, and it does.
When we exchange the identical ortho-LLE onto a frozen \texttt{distilbert-base-uncased} over 20 Newsgroups text (\hyperref[app:llm]{SI~\ref{app:llm}}: 5 in-distribution categories, 5 held-out near-OOD, word-shuffled far-OOD), all three structural findings surface. 
Orthonormality raises weight-space diversity $0.212\to0.909$ at no accuracy cost, the ortho-LLE outperforms the van-LLE on every scoring rule, near/far, and the scoring order is unchanged.
As on weak images, the unsupervised OC norm is weak (ROC AUC ca. 0.6) and the edge of the cov-LLE does not carry to these raw LM features, because the gain is gated by feature quality, reinforcing that it is the diversity mechanism, not the size of the gain from any one objective.
The findings also transfer up in scale (i.e., many classes). 
On a frozen pretrained 100-class CIFAR-100 backbone all three hold (ortho-LLE $>$ van-LLE, cov-LLE leads far-OOD, direction corrects OC on both near/far; all significant; \hyperref[app:cifar100]{SI~\ref{app:cifar100}}).

\section{Discussion}
\label{sec:discussion}
\noindent\textbf{Mitigating collapse in function space.}
A last-layer ensemble is an approach to turn one network into many, but its members, sharing every backbone gradient, can converge to a single function \citep{Kirsch::2025aa}.
Weight-orthonormality (OC) provides weight-space diversity inexpensively but reaches the predictions of the members only partially, and when placed on an intermediate embedding rather than the logit layer, not at all (\hyperref[sec:divobj]{Section~\ref{sec:divobj}}, inert on a two-layer head), an indirect correction strategy.
We instead mitigate the collapse directly in function space, with a covariance penalty on member activations (cov-LLE) that restores function-space diversity and, at matched $K$, recovers a substantial fraction of the prediction variance and calibration of a deep ensemble at $1\times$ cost (\autoref{tab:deep_ensemble_comparison}).
This gives the strongest of the OC and last-layer scores and significantly outperforms the weight-orthonormality of OC on far-OOD on two frozen backbones, confirming that last-layer diversity can recover the benefit of a full deep ensemble.
Viewing OC itself as a last-layer ensemble is what locates this correction, turning the equivalence into a design space in which weight-orthonormality is one \emph{indirect} instance of ``decorrelate the members'' and a direct approach offers an improvement.
The same view yields a scale-invariant direction score lifting the label-free near-OOD score above chance on every backbone, and an $\mathcal{O}(KC)$ or $\mathcal{O}(C)$ efficiency result over the $\mathcal{O}(K^2C)$ pairwise distribution metrics \citep{Gillis::2026aa}.

\smallskip\noindent\textbf{The structural foundation.}
Every detector here is $K$ orthonormal linear units on a frozen $\varphix$.
OC trains them to vanish on in-distribution data and scores their magnitude; the ortho-LLE trains them to classify and scores their disagreement.
The orthonormality defining OC is, in the ensemble, the correction for diversity collapse (restoring weight-space diversity at unchanged accuracy), so the two track each other on every backbone.
However, the scoring rule is coupled to the member objective: norm scoring requires unsupervised, disagreement scoring requires a supervised predictive distribution, and the \emph{supervised} $\times$ \emph{norm} combination is the failing corner that inverts.

\smallskip\noindent\textbf{Feature quality determines whether diversity converts to detection gains.}
Every detector here is a geometric score, so all share the feature-collapse bottleneck, but overcoming it requires more than making $\varphix$ bi-Lipschitz.
On strong CIFAR backbones, feature quality controls the label-free OC norm, spectral normalization is a minor supervised-side adjustment, and pushing conditioning via adversarial robustness is actively detrimental.
The OC norm needs strong features (\hyperref[sec:natural]{Section~\ref{sec:natural}}); the far-OOD advantage of the cov-LLE is feature-gated, decisive on strong frozen backbones (\hyperref[sec:divobj]{Section~\ref{sec:divobj}}) but absent on raw LM features and the weak CIFARNet.
Importantly, what is gated is the detection gain, not the collapse mitigation.
Cov-LLE restores function-space diversity and calibration regardless of backbone, but that diversity converts to a ROC AUC gain only when the features are strong.
With labels, a supervised ensemble scored by BALD still outperforms OC, which quantifies what labels add once the backbone is fixed.
Without labels, the label-free OC score is a usable single-forward-pass alternative once the direction score replaces the magnitude score (\hyperref[sec:direction]{Section~\ref{sec:direction}}), and it too is best on strong features.

\section{Conclusion}
\label{sec:conclusion}
A last-layer ensemble, $K$ linear units on one frozen feature map of which Orthonormal Certificates (OC) are the weight-orthonormal special case, is the cheapest single-pass route to ensemble uncertainty, but shared gradients can collapse its members onto a common function.
The weight-orthonormality of OC corrects this collapse only indirectly; a covariance penalty on member activations (cov-LLE) corrects it directly in function space, recovering much of the diversity and calibration of a deep ensemble at $1\times$ cost and giving the strongest of the OC and last-layer scores, although the feature-space baselines still lead on raw ROC AUC.
The same last-layer-ensemble view provides a scale-invariant direction score that repairs the label-free near-OOD failure.
Beneath all of it lies feature quality, not network conditioning, which decides whether the restored diversity becomes a detection gain.
A single label-free score best on both near- and far-OOD across all backbones remains open.

\bibliographystyle{acmart}
\bibliography{references}

@inproceedings{Amersfoort::2021aa,
	author = {van Amersfoort, Joost and Smith, Lewis and Jesson, Andrew and Key, Oscar and Gal, Yarin},
	booktitle = {NeurIPS workshop},
	publisher = {arXiv},
	address = {Online},
	title = {On feature collapse and deep kernel learning for single forward pass uncertainty},
	pages = {},
	year = {2021}}

@inproceedings{Amersfoort::2020aa,
	author = {van Amersfoort, Joost and Smith, Lewis and Teh, Yee Whye and Gal, Yarin},
	booktitle = {ICML},
	publisher = {PMLR},
	address = {Online},
	title = {Uncertainty estimation using a single deep deterministic neural network},
	pages = {},
	year = {2020}}

@inproceedings{Bardes::2022aa,
	author = {Bardes, A. and Ponce, J. and LeCun, Y.},
	booktitle = {ICLR},
	publisher = {OpenReview.net},
	address = {Online},
	title = {{VICReg}: {V}ariance-invariance-covariance Regularization for self-supervised learning},
	pages = {},
	year = {2022}}

@inproceedings{Croce::2021aa,
	author = {Croce, F. and Andriushchenko, M. and Sehwag, V. and Debenedetti, E. and Flammarion, N. and Chiang, M. and Mittal, P. and Hein, M.},
	booksubtitle = {Datasets and Benchmarks},
	booktitle = {NeurIPS},
	publisher = {Curran Associates Inc.},
	address = {Red Hook, NY, USA},
	title = {{RobustBench}: {A} standardized adversarial robustness benchmark},
	pages = {},
	year = {2021}}

@inproceedings{Daxberger::2021aa,
	author = {Daxberger, Erik and Kristiadi, Agustinus and Immer, Alexander and Eschenhagen, Runa and Bauer, Matthias and Hennig, Philipp},
	booktitle = {NeurIPS},
	publisher = {Curran Associates Inc.},
	address = {Red Hook, NY, USA},
	title = {Laplace redux -- effortless {B}ayesian deep learning},
	pages = {},
	year = {2021}}

@inproceedings{Gal::2016ab,
	author = {Gal, Yarin and Ghahramani, Zoubin},
	booktitle = {ICML},
	publisher = {PMLR},
	address = {Online},
	title = {Dropout as a {B}ayesian approximation: {R}epresenting model uncertainty in deep learning},
	pages = {},
	year = {2016}}

@inproceedings{Galdran::2023aa,
	author = {Galdran, Adrian and Verjans, Johan and Carneiro, Gustavo and Ballester, Miguel A. González},
	booktitle = {MICCAI},
	publisher = {Springer Nature Switzerland},
    address   = {Cham, Switzerland},
	title = {Multi-head multi-loss model calibration},
    doi = {10.1007/978-3-031-43898-1_11},
    volume = {14222},
	pages = {108--117},
	year = {2023}}

@article{Gillis::2026aa,
	author = {Gillis, H. Martin and Xu, Isaac and Trappenberg, Thomas},
	journal = {TMLR},
	volume = {6},
	title = {Variance-gated ensembles: {A}n epistemic-aware framework for uncertainty estimation},
	pages = {},
	year = {2026}}

@inproceedings{Harrison::2024aa,
	author = {Harrison, James and Willes, John and Snoek, Jasper},
	booktitle = {ICLR},
	publisher = {OpenReview.net},
	address = {Online},
	title = {Variational {B}ayesian last layers},
	pages = {},
	year = {2024}}

@inproceedings{Havasi::2021aa,
	author = {Havasi, Marton and Jenatton, Rodolphe and Fort, Stanislav and Liu, Jeremiah Zhe and Snoek, Jasper and Lakshminarayanan, Balaji and Dai, Andrew M. and Tran, Dustin},
	booktitle = {ICLR},
	publisher = {OpenReview.net},
	address = {Online},
	title = {Training independent subnetworks for robust prediction},
	pages = {},
	year = {2021}}

@inproceedings{Hendrycks::2017aa,
	author = {Hendrycks, Dan and Gimpel, Kevin},
	booktitle = {ICLR},
	publisher = {OpenReview.net},
	address = {Online},
	title = {A baseline for detecting misclassified and out-of-distribution examples in neural networks},
	pages = {},
	year = {2017}}

@misc{Houlsby::2011aa,
	author = {Houlsby, Neil and Huszár, Ferenc and Ghahramani, Zoubin and Lengyel, Máté},
	title = {{B}ayesian active learning for classification and preference learning},
	howpublished = {arXiv preprint},
	doi = {10.48550/arXiv.1112.5745},
	pages = {},
	year = {2011}}

@article{Kirsch::2025aa,
	author = {Kirsch, Andreas},
	doi = {10.48550/arXiv.2409.02628},
	journal = {TMLR},
	volume = {5},
	pages = {1--16},
	title = {(implicit) ensembles of ensembles: {E}pistemic uncertainty collapse in large models},
	year = {2025}}

@inproceedings{Kornblith::2019aa,
	author = {Kornblith, S. and Norouzi, M. and Lee, H. and Hinton, G.},
	booktitle = {ICML},
	publisher = {PMLR},
	address = {Online},
	title = {Similarity of neural network representations revisited},
	pages = {},
	year = {2019}}

@inproceedings{Kristiadi::2020aa,
	author = {Kristiadi, Agustinus and Hein, Matthias and Hennig, Philipp},
	booktitle = {ICML},
	doi = {10.48550/arXiv.2002.10118},
	publisher = {PMLR},
	address = {Online},
	title = {Being {B}ayesian, even just a bit, fixes overconfidence in {ReLU} networks},
	pages = {},
	year = {2020}}

@inproceedings{Lakshminarayanan::2017aa,
	author = {Lakshminarayanan, Balaji and Pritzel, Alexander and Blundell, Charles},
	booktitle = {NeurIPS},
	publisher = {Curran Associates Inc.},
	address = {Red Hook, NY, USA},
	title = {Simple and scalable predictive uncertainty estimation using deep ensembles},
	pages = {},
	year = {2017}}

@misc{Lee::2015aa,
	author = {Lee, Stefan and Purushwalkam, Senthil and Cogswell, Michael and Crandall, David and Batra, Dhruv},
	doi = {10.48550/arXiv.1511.06314},
	howpublished = {arXiv preprint},
	title = {Why {M} heads are better than one: {T}raining a diverse ensemble of deep networks},
	pages = {},
	year = {2015}}

@inproceedings{Lee::2018ac,
	author = {Lee, Kimin and Lee, Kibok and Lee, Honglak and Shin, Jinwoo},
	publisher = {Curran Associates Inc.},
	address = {Red Hook, NY, USA},
	title = {A simple unified framework for detecting out-of-distribution samples and adversarial attacks},
	booktitle = {NeurIPS},
	pages = {},
	year = {2018}}

@inproceedings{Liang::2018aa,
	author = {Liang, Shiyu and Li, Yixuan and Srikant, R.},
	booktitle = {ICLR},
	publisher = {OpenReview.net},
	address = {Online},
	title = {Enhancing the reliability of out-of-distribution image detection in neural networks},
	pages = {},
	year = {2018}}

@article{Liu::2022ad,
	author = {Liu, Jeremiah Zhe and Padhy, Shreyas and Ren, Jie and Lin, Zi and Wen, Yeming and Jerfel, Ghassen and Nado, Zachary and Snoek, Jasper and Tran, Dustin and Lakshminarayanan, Balaji},
	journal = {JMLR},
	title = {A simple approach to improve single-model deep uncertainty via distance-awareness},
	doi = {10.5555/3648699.3648741},
	volume = {23},
	pages = {1--63},
	year = {2022}}

@inproceedings{Liu::2020ab,
	author = {Liu, Weitang and Wang, Xiaoyun and Owens, John D. and Li, Yixuan},
	booktitle = {NeurIPS},
	publisher = {Curran Associates Inc.},
	address = {Red Hook, NY, USA},
	title = {Energy-based out-of-distribution detection},
	pages = {},
	year = {2020}}

@inproceedings{Miyato::2018aa,
	author = {Miyato, Takeru and Kataoka, Toshiki and Koyama, Masanori and Yoshida, Yuichi},
	booktitle = {ICLR},
	title = {Spectral normalization for generative adversarial networks},
	publisher = {OpenReview.net},
	address = {Online},
	pages = {},
	year = {2018}}

@inproceedings{Nalisnick::2019aa,
	author = {Nalisnick, E. and Matsukawa, A. and Teh, Y. W. and Gorur, D. and Lakshminarayanan, B.},
	booktitle = {ICLR},
	publisher = {OpenReview.net},
	address = {Online},
	title = {Do deep generative models know what they don't know?},
	pages = {},
	year = {2019}}

@inproceedings{Osband::2023aa,
	author = {Osband, Ian and Wen, Zheng and Asghari, Seyed Mohammad and Dwaracherla, Vikranth and Ibrahimi, Morteza and Lu, Xiuyuan and Van Roy, Benjamin},
	booktitle = {NeurIPS},
	publisher = {Curran Associates Inc.},
	address = {Red Hook, NY, USA},
	title = {Epistemic neural networks},
	pages = {},
	year = {2023}}

@inproceedings{Sun::2022aa,
	author = {Sun, Y. and Ming, Y. and Zhu, X. and Li, Y.},
	booktitle = {ICML},
	publisher = {PMLR},
	address = {Online},
	title = {Out-of-distribution detection with deep nearest neighbors},
	pages = {},
	year = {2022}}

@inproceedings{Sun::2021aa,
	author = {Sun, Y. and Guo, C. and Li, Y.},
	booktitle = {NeurIPS},
	publisher = {Curran Associates Inc.},
	address = {Red Hook, NY, USA},
	title = {{ReAct}: {O}ut-of-distribution detection with rectified activations},
	pages = {},
	year = {2021}}

@inproceedings{Tagasovska::2019ab,
	author = {Tagasovska, Natasa and Lopez-Paz, David},
	booktitle = {NeurIPS},
	publisher = {Curran Associates Inc.},
	address = {Red Hook, NY, USA},
	title = {Single-Model Uncertainties for Deep Learning},
	pages = {},
	year = {2019}}

@inproceedings{Wang::2022aa,
	author = {Wang, H. and Li, Z. and Feng, L. and Zhang, W.},
	booktitle = {CVPR},
	publisher = {IEEE},
	address = {Los Alamitos, CA, USA},
	title = {{ViM}: {O}ut-of-distribution with virtual-logit matching},
	pages = {},
	year = {2022}}

@inproceedings{Wen::2020aa,
	author = {Wen, Yeming and Tran, Dustin and Ba, Jimmy},
	booktitle = {ICLR},
	publisher = {OpenReview.net},
	address = {Online},
	title = {{BatchEnsemble}: {A}n alternative approach to efficient ensemble and lifelong learning},
	pages = {},
	year = {2020}}

@article{Yang::2024aa,
	author = {Yang, J. and Zhou, K. and Li, Y. and Liu, Z.},
	doi = {10.1007/s11263-024-02117-4},
	journal = {IJCV},
	title = {Generalized out-of-distribution detection: {A} survey},
	volume = {132},
	pages = {5635--5662},
	year = {2024}}

@inproceedings{Yang::2022aa,
	author = {Yang, Jingkang and Wang, Pengyun and Zou, Dejian and Zhou, Zitang and Ding, Kunyuan and Peng, Wenxuan and Wang, Haoqi and Chen, Guangyao and Li, Bo and Sun, Yiyou and Du, Xuefeng and Zhou, Kaiyang and Zhang, Wayne and Hendrycks, Dan and Li, Yixuan and Liu, Ziwei},
	title = {{OpenOOD}: {B}enchmarking generalized out-of-distribution detection},
	booktitle = {NeurIPS},
	publisher = {Curran Associates Inc.},
	address = {Red Hook, NY, USA},
	pages = {1--14},
	year = {2022}}

@inproceedings{Zbontar::2021aa,
	author = {Zbontar, Jure and Jing, Li and Misra, Ishan and {LeCun}, Yann and Deny, Stéphane},
	title = {{Barlow Twins}: {S}elf-supervised learning via redundancy reduction},
	booktitle = {ICML},
	publisher = {PMLR},
	address = {Online},
	pages = {1--11},
	year = {2021}}

\appendix
\renewcommand{\thesection}{S\arabic{section}}
\renewcommand{\theequation}{S\arabic{equation}}
\renewcommand{\thefigure}{S\arabic{figure}}
\renewcommand{\thetable}{S\arabic{table}}
\renewcommand{\thepage}{S\arabic{page}}
\setcounter{section}{0}
\setcounter{equation}{0}
\setcounter{figure}{0}
\setcounter{table}{0}
\setcounter{page}{1}

\section*{Supporting Information}
%
\section{Comparison against post-hoc baselines}
\label{app:openood}
\begin{figure}[h]
\centering
\includegraphics[width=\columnwidth]{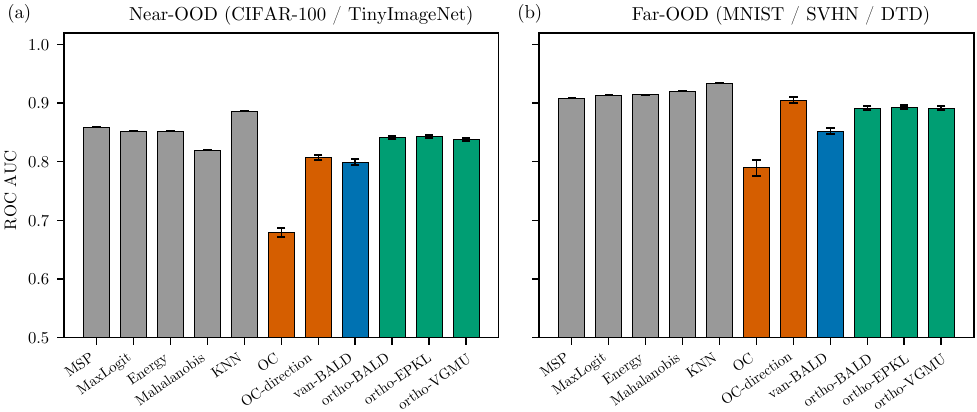}
\caption{%
    \textbf{The diversity correction replicates (ortho-LLE $>$ van-LLE) and the direction score lifts the label-free OC.} CIFAR-10 ROC AUC per detector on a frozen ResNet-18 (OpenOOD protocol), near- and far-OOD.
}
\label{fig:openood_comparison}
\end{figure}

We reproduce the OpenOOD \citep{Yang::2022aa} CIFAR-10 protocol so that every detector runs on the same frozen ResNet-18 (CIFAR-10, accuracy: $0.940$). 
ID $=$ CIFAR-10; near-OOD $=\{$CIFAR-100, TinyImageNet$\}$; far-OOD $=\{$MNIST, SVHN, DTD$\}$ (Places365 omitted); metrics are ROC AUC and FPR@95 averaged within each group. 
Every baseline uses its standard parameter-free or OpenOOD-default setting with no per-OOD-set tuning. 
MSP and energy are parameter-free (max-softmax; logsumexp logits at temperature~$1$), Mahalanobis fits a single class-conditional Gaussian with tied covariance ($10^{-4}$ shrinkage) on the penultimate features, and KNN uses the $k{=}50$ $k$-th nearest-neighbor distance on $L_2$-normalized features \citep{Sun::2022aa}. 
All detectors use the same frozen features, and where a training reference is required (the reference features of KNN and the class statistics of Mahalanobis), it is the full CIFAR-10 training split ($N{=}50{,}000$ images); no method sees the OOD data. 
Hyperparameters are thus held fixed across methods, so the comparison isolates the detector rather than its tuning. 
All main-text OpenOOD numbers (\autoref{tab:ood_baselines}, \autoref{fig:openood_comparison}) come from this setup. 
The diversity repair replicates here (weight-space diversity $0.273\to0.991$), the ortho-LLE and cov-LLE gains of \hyperref[sec:divobj]{Section~\ref{sec:divobj}} hold, and KNN still leads on raw ROC AUC.

\section{A function-space diversity objective}
\label{app:divobj}
This section provides the detailed MNIST frozen/joint breakdown supporting the main-text result (\hyperref[sec:divobj]{Section~\ref{sec:divobj}}, where the covariance objective is defined, \autoref{eq:cov}, and shown to outperform the strongest weight-space ensemble on strong CIFAR backbones).
The orthonormality penalty acts on the weights of the heads; the alternative acts on their activations, decorrelating the embeddings of the heads on data, a Barlow-Twins / VICReg-style \citep{Zbontar::2021aa,Bardes::2022aa} covariance penalty $\lVert \mathrm{cov}(Z)-I\rVert^2$ on the centered, concatenated per-head embeddings $Z$.
We compare the two at the study default $K=32$ ($\lambda_\mathrm{cov}{=}0.5$), with a small per-head embedding ($\varphi\to z_k\in\mathbb{R}^{32}\to\text{logits}_k$) on MNIST 0--4 over 10 seeds.
Function-space diversity is scored by $\delta_{\mathrm{f}}$ ($1{-}$linear CKA \citep{Kornblith::2019aa}, a bounded $[0,1]$ measure invariant to invertible linear reparametrization of the embedding, so it registers decorrelation the classifier layer above would otherwise negate) and member prediction variance (\autoref{tab:covlle_mnist}).
\begin{table*}[h]
\centering
\caption{%
    \textbf{On a two-layer head, weight-orthonormality is functionally inert while the covariance penalty increases near-OOD detection.} Diversity objectives at $\bm{K{=}32}$ on MNIST 0--4 (van-LLE, ortho-LLE, cov-LLE at $\bm{\lambda_\mathrm{cov}{=}0.5}$), frozen and jointly-trained $\bm{\varphi}$.\textsuperscript{a}
    }
\label{tab:covlle_mnist}
\footnotesize
\begin{tabular*}{\textwidth}{@{\extracolsep{\fill}}lccc@{}}
\toprule
\textbf{Metric} 
& \textbf{van-LLE} 
& \textbf{ortho-LLE} 
& \textbf{cov-LLE} 
\\
\midrule
\multicolumn{4}{l}{\emph{frozen $\varphi$}} \\
function-space diversity $\delta_{\mathrm{f}}$ ($1{-}$CKA) & $0.015\pm0.005$ & $0.015\pm0.005$ & $\bm{0.520\pm0.247}$ \\
prediction variance ($\times10^{-3}$) & $0.124\pm0.042$ & $0.117\pm0.047$ & $\bm{0.722\pm0.815}$ \\
near-OOD (ROC AUC)  & $0.888\pm0.008$ & $0.887\pm0.009$ & $\bm{0.944\pm0.022}$ \\
far-OOD (ROC AUC) & $0.980\pm0.006$ & $0.980\pm0.007$ & $\bm{0.983\pm0.010}$ \\
in-distribution ECE & $\bm{0.002\pm0.000}$ & $\bm{0.002\pm0.000}$ & $0.010\pm0.007$ \\
accuracy & $0.998\pm0.000$ & $0.998\pm0.000$ & $0.998\pm0.001$ \\
\midrule
\multicolumn{4}{l}{\emph{jointly-trained $\varphi$}} \\
function-space diversity $\delta_{\mathrm{f}}$ ($1{-}$CKA) & $0.024\pm0.002$ & $0.024\pm0.002$ & $\bm{0.113\pm0.009}$ \\
prediction variance ($\times10^{-3}$) & $0.023\pm0.007$ & $0.024\pm0.003$ & $\bm{0.058\pm0.006}$ \\
near-OOD (ROC AUC) & $0.916\pm0.027$ & $0.916\pm0.026$ & $\bm{0.969\pm0.004}$ \\
far-OOD (ROC AUC) & $0.983\pm0.005$ & $0.982\pm0.007$ & $\bm{0.992\pm0.003}$ \\
in-distribution ECE & $\bm{0.002\pm0.000}$ & $\bm{0.002\pm0.000}$ & $0.061\pm0.004$ \\
accuracy & $0.997\pm0.001$ & $0.997\pm0.001$ & $0.999\pm0.000$ \\
\bottomrule
\end{tabular*}
\vspace{2pt}
\noindent\parbox{\textwidth}{%
    \footnotesize\textsuperscript{a}
    Function-space diversity $=$ member prediction variance and $\delta_{\mathrm{f}}=1-$ mean pairwise linear CKA (higher $=$ more diverse).
    \textbf{Bold} $=$ best method per row (higher $\delta_{\mathrm{f}}$, prediction variance, and ROC AUC; lower ECE).
}
\end{table*}

The contrast is sharp (\autoref{tab:covlle_mnist}):
(i) In this architecture, weight-orthonormality is nearly functionally inert, its frozen function-space diversity and prediction variance indistinguishable from vanilla.
This is not a contradiction of the main results.
There the penalty acts on the single layer that directly produces the logits, so it raises weight-space diversity and OOD detection; here it acts on the embedding map and is negated by the classifier layer above it, leaving the predictions unchanged, which is exactly why acting on the activations is the more direct intervention.
The covariance objective instead decorrelates the heads.
(ii) That diversity helps where detection is hard, the frozen near-OOD ROC AUC rising at no accuracy cost.
(iii) At the detection-optimal $\lambda_\mathrm{cov}{=}0.5$, selected by a MNIST $\lambda_\mathrm{cov}$ ablation that lands on the same value as the CIFAR ablation (\hyperref[sec:divobj]{Section~\ref{sec:divobj}}, \autoref{fig:covlle_lambda_sweep}), this incurs no meaningful accuracy or calibration penalty.
Far-OOD stays saturated (ca. $0.98$), the frozen ECE barely moves ($0.002\to0.010$), and the jointly-trained $\varphi$ no longer over-decorrelates (near-OOD $0.969$ and far-OOD $0.992$, both best, with only a small ECE rise to $0.061$).
Increasing $\lambda_\mathrm{cov}$ higher results in negligible near-OOD, but steadily degrades calibration on both datasets, which is why we fix $\lambda_\mathrm{cov}{=}0.5$ throughout.

\smallskip\noindent\textbf{Number of units $\bm{K}$.}
The ensembles are robust to $K$ (\autoref{tab:mnist_k_ablation}; ortho even improves slightly).
OC has an optimum at $K\!\approx\!d/4=32$ and collapses as $K\to d$: at $K=128=d$ it inverts to $0.139/0.044$ (near/far).
Requiring as many orthonormal directions that vanish on ID data as the feature dimension is impossible. 
The ID data spans the full space, so the certificates cannot all vanish on it. 
This is a real constraint on OC that the supervised ensemble, a full classifier per member, does not share, since its members are not required to vanish.
We therefore fix $K=32$ throughout the study.
It is the MNIST optimum of OC ($d/4$), a safe choice on CIFAR (where the supervised ensembles are essentially $K$-insensitive), and it resides comfortably below the $K{=}d$ collapse on every backbone ($d\in\{128,256,512\}$).
The only deliberate exceptions are this ablation itself and the deep-ensemble comparison (\autoref{tab:deep_ensemble_comparison}), fixed at the matched $K{=}10$ of its $K$ independently-trained networks.
\begin{table*}[h]
\centering
\caption{%
     \textbf{OC has an optimum at $\bm{K{=}d/4}$ and collapses as $\bm{K{\to}d}$; the supervised ensembles are essentially $\bm{K}$-insensitive.} ROC AUC near-/far-OOD vs.\ number of units $\bm{K}$ on MNIST 0--4.\textsuperscript{a}
}
\label{tab:mnist_k_ablation}
\footnotesize
\begin{tabular*}{\textwidth}{@{\extracolsep{\fill}}l*{6}{c}@{}}
\toprule
 & \multicolumn{2}{c}{\textbf{OC}} & \multicolumn{2}{c}{\textbf{van-LLE}} & \multicolumn{2}{c}{\textbf{ortho-LLE}} \\
\cmidrule(lr){2-3}\cmidrule(lr){4-5}\cmidrule(lr){6-7}
$K$ & near & far & near & far & near & far \\
\midrule
4   & $0.762\pm0.034$ & $0.802\pm0.080$ & $0.924\pm0.007$ & $0.990\pm0.002$ & $0.928\pm0.007$ & $0.991\pm0.002$ \\
8   & $0.815\pm0.022$ & $0.843\pm0.033$ & $0.926\pm0.006$ & $0.991\pm0.001$ & $0.931\pm0.007$ & $0.992\pm0.001$ \\
16  & $0.851\pm0.021$ & $0.860\pm0.042$ & $0.926\pm0.006$ & $0.991\pm0.001$ & $0.933\pm0.007$ & $0.993\pm0.001$ \\
32  & $\bm{0.866\pm0.026}$ & $\bm{0.889\pm0.035}$ & $0.925\pm0.005$ & $0.991\pm0.001$ & $0.935\pm0.007$ & $0.993\pm0.001$ \\
64  & $0.839\pm0.032$ & $0.850\pm0.038$ & $0.923\pm0.006$ & $0.991\pm0.002$ & $0.936\pm0.007$ & $0.994\pm0.001$ \\
128 & $\mathit{0.139\pm0.037}$ & $\mathit{0.044\pm0.025}$ & $0.923\pm0.006$ & $0.991\pm0.002$ & $\bm{0.939\pm0.007}$ & $\bm{0.994\pm0.001}$ \\
\bottomrule
\end{tabular*}
\vspace{2pt}
\noindent\parbox{\textwidth}{%
    \footnotesize\textsuperscript{a}
    \textbf{Bold} $=$ OC optimum ($K{=}32$), best ortho ($K{=}128$); \emph{italic} $=$ OC collapse at $K{=}128$.
}
\end{table*}

\section{The direction score, additional results}
\label{app:direction}
\begin{figure}[h]
\centering
\includegraphics[width=\columnwidth]{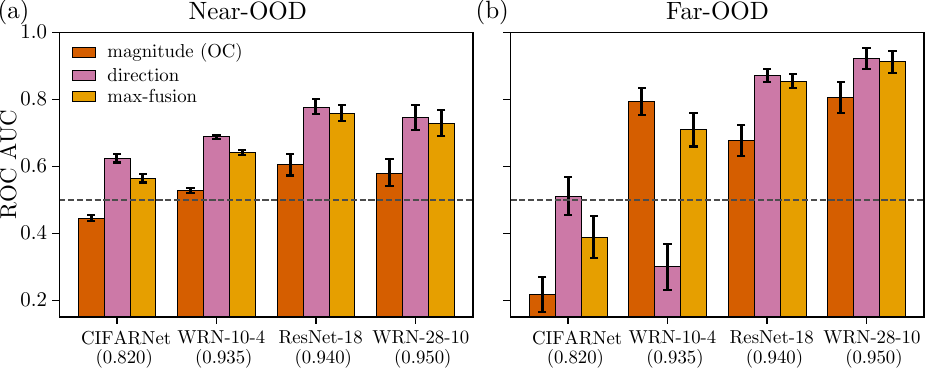}
\caption{%
    A direction score of the same certificates repairs near-OOD on every backbone; magnitude and direction are complementary on far-OOD. Magnitude, direction, and their max-fusion across four frozen backbones.
}
\label{fig:oc_direction}
\end{figure}

Here, we fit OC exactly as in the main text ($K{=}32$, $\lambda_\mathrm{oc}{=}5$, standardized features) and only change the score, on four frozen backbones spanning the study (\autoref{tab:direction_score}, \autoref{fig:oc_direction}; the certificates are re-fit independently, so the magnitude baselines differ slightly from the main-text tables).

The four backbones use their native protocols: CIFARNet and WRN-10-4 with CIFAR-10 0--4 ID / 5--9 near / SVHN far; ResNet-18 with CIFAR-10 ID / \{CIFAR-100, TinyImageNet\} near / \{SVHN, MNIST, DTD\} far; and the standard WRN-28-10 with CIFAR-10 ID / CIFAR-100 near / SVHN far. 
The direction score behaves differently on near- and far-OOD:
(i) Near-OOD. 
The direction score raises near-OOD on every backbone (CIFARNet $0.446\!\to\!0.624$, WRN-10-4 $0.529\!\to\!0.689$, ResNet-18 $0.605\!\to\!0.778$, WRN-28-10 $0.581\!\to\!0.746$), all significant.
This overturns the near-OOD conclusion of the main text and closes much of the gap to the supervised ensembles.
(ii) Far-OOD. 
The direction score de-inverts three of the four backbones (\autoref{tab:direction_score}; the values are in the main text).
However, on WRN-10-4 (whose main-text ``recovery'' detects SVHN precisely by its certificate magnitude), dividing that magnitude out significantly inverts the score ($0.794\!\to\!0.300$, $t\approx-24$). 
Magnitude thus identifies scale-anomalous far-OOD, and direction identifies the rest.
We therefore combine the two by max-fusion: standardize each score with training-set statistics and take the elementwise maximum
\begin{equation}
    \label{eq:maxfusion}
    \mathrm{S_{max}}(x) = 
    \max\!\left(
    \frac{\mathrm{S_N}(x) - \mu_{\mathrm{N}}}{\sigma_{\mathrm{N}}},\;
    \frac{\mathrm{S_D}(x) - \mu_{\mathrm{D}}}{\sigma_{\mathrm{D}}}
    \right),
\end{equation}
where $\mu_{\mathrm{N}},\sigma_{\mathrm{N}}$ and $\mu_{\mathrm{D}},\sigma_{\mathrm{D}}$ are the mean and standard deviation of $\mathrm{S_N}$ and $\mathrm{S_D}$ over the training (ID) set.
This is label-free and, being a max of two standardized scores, never inverts; it recovers most of both on the conflicting backbone (WRN-10-4 near $0.642$, far $0.710$), a safe default, although it does not match the best single score on near- or far-OOD.

\section{Robustness ablation at fixed architecture}
\label{app:robustbench}
\begin{figure}[h]
\centering
\includegraphics[width=\columnwidth]{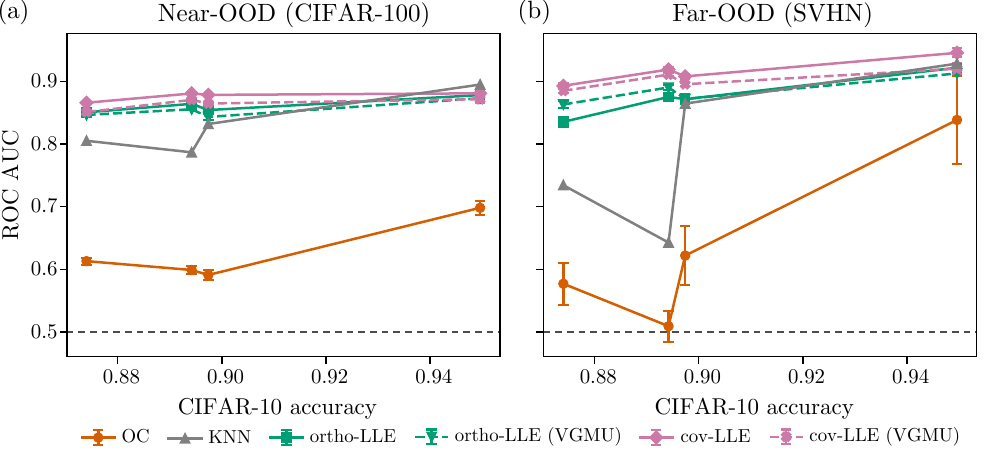}
\caption{%
    \textbf{Detection degrades as backbones get more robust.} ROC AUC for each detector on the four WRN-28-10 checkpoints with respect to CIFAR-10 accuracy (three robust checkpoints; standard at right). cov-LLE is best on both near- and far-OOD, OC weakest.
}
\label{fig:robustbench_sweep}
\end{figure}

To vary the smoothness (Lipschitz conditioning) of $\varphi$ while holding architecture fixed, we take four WRN-28-10 CIFAR-10 checkpoints from RobustBench \citep{Croce::2021aa}, one standard and three $\ell_\infty$-adversarially trained (adversarial training being a strong form of Lipschitz/smooth\-ness conditioning), freeze each, and evaluate OC, the ortho-LLE, and the cov-LLE off the penultimate features (ID $=$ CIFAR-10, near $=$ CIFAR-100, far $=$ SVH).
The result is unambiguous (\autoref{tab:robustbench}, \autoref{fig:robustbench_sweep}). 
\begin{table*}[h]
\centering
\caption{%
    \textbf{Adversarial robustness hurts OOD detection.} OC, ortho-LLE, and cov-LLE on four frozen RobustBench WRN-28-10 checkpoints (CIFAR-10 ID; one standard, three $\ell_\infty$-robust). Every detector is significantly worse on the robust backbones; cov-LLE leads throughout.\textsuperscript{a}
}
\label{tab:robustbench}
\footnotesize
\begin{tabular*}{\textwidth}{@{\extracolsep{\fill}}llcccccccc@{}}
\toprule
& & \multicolumn{4}{c}{\textbf{Near-OOD (CIFAR-100)}} & \multicolumn{4}{c}{\textbf{Far-OOD (SVHN)}} \\
\cmidrule(lr){3-6}\cmidrule(lr){7-10}
\textbf{Detector} & \textbf{Score} & Standard & Carmon & Gowal & Rebuffi & Standard & Carmon & Gowal & Rebuffi \\
\midrule
\multicolumn{2}{@{}l}{OC} & $\bm{0.698\pm0.011}$ & $0.591\pm0.008$ & $0.599\pm0.006$ & $0.613\pm0.005$ & $\bm{0.839\pm0.070}$ & $0.622\pm0.047$ & $0.509\pm0.025$ & $0.577\pm0.034$ \\
\midrule
\multirow{3}{*}{ortho-LLE} & BALD & $\bm{0.878\pm0.001}$ & $0.854\pm0.005$ & $0.864\pm0.002$ & $0.851\pm0.003$ & $\bm{0.922\pm0.004}$ & $0.872\pm0.008$ & $0.875\pm0.006$ & $0.836\pm0.008$ \\
 & EPKL & $\bm{0.878\pm0.001}$ & $0.855\pm0.005$ & $0.862\pm0.002$ & $0.847\pm0.003$ & $\bm{0.922\pm0.004}$ & $0.871\pm0.008$ & $0.864\pm0.006$ & $0.827\pm0.008$ \\
 & VGMU & $\bm{0.873\pm0.000}$ & $0.843\pm0.005$ & $0.856\pm0.002$ & $0.847\pm0.003$ & $\bm{0.913\pm0.002}$ & $0.872\pm0.008$ & $0.891\pm0.005$ & $0.863\pm0.006$ \\
\midrule
\multirow{3}{*}{cov-LLE} & BALD & $0.881\pm0.005$ & $0.879\pm0.001$ & $0.881\pm0.001$ & $0.866\pm0.002$ & $\bm{0.946\pm0.007}$ & $0.908\pm0.004$ & $0.919\pm0.003$ & $0.893\pm0.005$ \\
 & EPKL & $0.881\pm0.005$ & $0.879\pm0.001$ & $0.880\pm0.001$ & $0.867\pm0.002$ & $\bm{0.947\pm0.007}$ & $0.908\pm0.004$ & $0.916\pm0.003$ & $0.894\pm0.005$ \\
 & VGMU & $0.872\pm0.006$ & $0.865\pm0.001$ & $0.871\pm0.001$ & $0.851\pm0.002$ & $\bm{0.921\pm0.009}$ & $0.896\pm0.002$ & $0.911\pm0.002$ & $0.885\pm0.003$ \\
\bottomrule
\end{tabular*}
\vspace{2pt}
\noindent\parbox{\textwidth}{%
    \footnotesize\textsuperscript{a}
    Near-OOD $=$ CIFAR-100, far-OOD $=$ SVHN. ROC AUC, 10 seeds. 
    Clean accuracy: standard $0.950$, Carmon $0.897$, Gowal $0.894$, Rebuffi $0.874$ (robust checkpoints). 
    \textbf{Bold} $=$ significantly-best backbone per detector, near- and far-OOD (paired $t$, $p<0.05$, $n=10$), comparing backbones within a row rather than detectors; cov-LLE near-OOD ties the strongest robust checkpoints.
}
\end{table*}

The robust backbones are worse for OOD detection (paired $t$, $p<0.05$; the only exception is cov-LLE near-OOD, where the strongest robust checkpoints match the standard backbone), and the loss in detection follows the decrease in clean accuracy.
The ordering and per-checkpoint values match \hyperref[sec:featquality]{Section~\ref{sec:featquality}} and \autoref{tab:robustbench}, where the label-free OC norm collapses toward chance, the supervised ortho-LLE is more resilient, and the function-space cov-LLE is the best of the three on every checkpoint.
It too degrades with robustness; even KNN declines (\autoref{fig:robustbench_sweep}). 
So conditioning-via-robustness is not the determining factor; feature quality is.
Adversarial training also lowers clean accuracy and reshapes the feature geometry, so this conflates smoothness with feature quality; it does not isolate the spectral-norm operator itself, which we do in the next section.

\section{Spectral-normalization ablation}
\label{app:snablation}
\begin{table*}[ht]
\centering
\caption{%
    \textbf{Spectral normalization helps the supervised ensembles on far-OOD but leaves the label-free OC norm inverted, so smoothness is neither sufficient nor necessary for it.} ROC AUC for the label-free OC norm and for the ortho-LLE and cov-LLE on the CIFARNet at three conditioning levels (plain, $+$SN, $+$SN$+$residual).\textsuperscript{a}
    }
\label{tab:cifar_spectralnorm_ablation}
\footnotesize
\begin{tabular*}{\textwidth}{@{\extracolsep{\fill}}ll*{3}{c}*{3}{c}@{}}
\toprule
& & \multicolumn{3}{c}{\textbf{Near-OOD}} & \multicolumn{3}{c}{\textbf{Far-OOD}} \\
\cmidrule(lr){3-5}\cmidrule(lr){6-8}
\textbf{Detector} & \textbf{Score} & plain & $+$SN & $+$SN$+$residual & plain & $+$SN & $+$SN$+$residual \\
\midrule
\multicolumn{2}{@{}l}{OC} & $0.456\pm0.013$ & $0.473\pm0.014$ & $\bm{0.505\pm0.009}$ & $0.305\pm0.067$ & $0.315\pm0.049$ & $0.202\pm0.028$ \\
\midrule
\multirow{3}{*}{ortho-LLE} & BALD & $0.637\pm0.012$ & $0.638\pm0.012$ & $\bm{0.659\pm0.007}$ & $0.676\pm0.035$ & $0.641\pm0.042$ & $\bm{0.737\pm0.020}$ \\
 & EPKL & $0.638\pm0.012$ & $0.638\pm0.012$ & $\bm{0.661\pm0.007}$ & $0.677\pm0.035$ & $0.641\pm0.042$ & $\bm{0.738\pm0.020}$ \\
 & VGMU & $0.630\pm0.012$ & $0.626\pm0.013$ & $\bm{0.651\pm0.007}$ & $0.709\pm0.038$ & $0.703\pm0.024$ & $\bm{0.770\pm0.023}$ \\
\midrule
\multirow{3}{*}{cov-LLE} & BALD & $0.577\pm0.016$ & $0.593\pm0.015$ & $\bm{0.632\pm0.015}$ & $0.544\pm0.033$ & $0.576\pm0.075$ & $\bm{0.628\pm0.039}$ \\
 & EPKL & $0.574\pm0.017$ & $0.591\pm0.015$ & $\bm{0.629\pm0.016}$ & $0.538\pm0.032$ & $0.572\pm0.075$ & $\bm{0.623\pm0.039}$ \\
 & VGMU & $0.595\pm0.014$ & $0.612\pm0.013$ & $\bm{0.635\pm0.014}$ & $0.601\pm0.039$ & $0.651\pm0.061$ & $\bm{0.670\pm0.042}$ \\
\bottomrule
\end{tabular*}
\vspace{2pt}
\noindent\parbox{\textwidth}{%
    \footnotesize\textsuperscript{a}
    ROC AUC; near-OOD $=$ CIFAR-10 5--9, far-OOD $=$ SVHN. 
    Accuracy: plain $0.802$, $+$SN $0.811$, $+$SN$+$residual $0.840$. 
    \textbf{Bold} $=$ best backbone per detector$\times$OOD.
}
\end{table*}

The RobustBench probe (\hyperref[app:robustbench]{SI~\ref{app:robustbench}}) conflates Lipschitz conditioning with feature quality, since adversarial training also lowers clean accuracy. 
To isolate the spectral-norm operator itself, we hold the CIFARNet architecture fixed and add conditioning in two steps: $+$SN and $+$SN$+$residual, retraining each and scoring OC, the ortho-LLE, and the cov-LLE off the frozen penultimate features (\autoref{tab:cifar_spectralnorm_ablation}; ID $=$ CIFAR-10 0--4, near $=$ 5--9, far $=$ SVHN).
Spectral normalization plus a residual lifts the supervised ensembles on far-OOD and modestly on near-OOD (\autoref{tab:cifar_spectralnorm_ablation}; the cov-LLE far-OOD BALD rises $0.544\to0.628$), but the label-free OC norm stays inverted on far-OOD and only reaches chance on near-OOD ($0.456\to0.505$).
Smoothness helps the supervised ensembles but does not repair the label-free norm, which does not need it either, matching the RobustBench conclusion that clean feature quality, not conditioning, controls the label-free score.

\section{The ortho-LLE on a frozen LLM}
\label{app:llm}
As a demonstration of modality generality we exchange the identical ortho-LLE (same $K=32$, $\lambda_\mathrm{ortho}=3$) onto a frozen \texttt{distilbert\-base-uncased}: its mean-pooled last hidden state is $\varphi$ (768-d), the heads classify 5 of the 20 Newsgroups categories, 5 related categories are held out as near-OOD, and word-shuffled documents serve as a far-OOD analog.
All three carry over:
(i) Orthonormality raises inter-member weight-space diversity at no accuracy cost ($0.872\to0.879$).
(ii) The ortho-LLE outperforms the van-LLE on every scoring rule, largest on far-OOD and the disagreement scores (BALD far $0.825\to0.881$, $t\approx14$).
(iii) The score ordering is unchanged, disagreement (BALD/EPKL) ranking highest and the margin/confidence scores lowest.
Consistent with the feature-quality hypothesis, the unsupervised OC norm is weak, as expected for an off-the-shelf LM that is not bi-Lipschitz-regularized, and the function-space cov-LLE does not extend its image-domain advantage to these raw LM features (it ties the ortho-LLE on far-OOD, $0.881$ vs.\ $0.886$, within noise, but is significantly worse on near-OOD, $0.773\to0.752$, $t\approx-14$).
So the diversity mechanism transfers across modality, while the size of the gain from any one objective remains feature-quality-gated.

\section{A pretrained 100-class CIFAR-100 backbone}
\label{app:cifar100}
To test the findings at a higher class count, we repeat the OpenOOD-style comparison on a frozen, pretrained CIFAR-100 classifier (a standard VGG; plain cross-entropy, non-robust; accuracy $0.740$; penultimate $d{=}512$, so $K{=}32$ sits in the same $d/16$ regime as the ResNet-18 study), with ID $=$ CIFAR-100, near $=$ CIFAR-10 / TinyImageNet, far $=$ SVHN / DTD / MNIST. 
All three headline findings survive on this harder, un-tuned setting:
(i) Orthonormality still corrects the weight-space diversity ($0.210\to0.973$) and the ortho-LLE significantly outperforms the van-LLE (near $0.743\to0.756$, $t\approx10$; far $0.715\to0.745$, $t\approx5.2$). 
(ii) The function-space cov-LLE significantly outperforms the ortho-LLE on far-OOD ($0.745\to0.761$, $t\approx2.5$) and ties it on near, exactly the cross-backbone pattern of \hyperref[sec:divobj]{Section~\ref{sec:divobj}}, where the far-OOD gain is the robust one.
(iii) The direction score significantly repairs the OC norm on both near- and far-OOD (near $0.747\to0.765$, $t\approx65$; far $0.710\to0.768$, $t\approx22$), raising it to parity with the post-hoc baselines. 
Differences are small (CIFAR-100 is harder; every detector lands in $0.74$--$0.78$ ROC AUC), but the mechanism transfers intact to a pretrained backbone.
\end{document}